%% file: neurips_2023.tex
\documentclass{article}

   \PassOptionsToPackage{numbers, compress}{natbib}

    \usepackage[final]{neurips_2023}

\usepackage[utf8]{inputenc} 
\usepackage[T1]{fontenc}    
\usepackage{hyperref}       
\usepackage{url}            
\usepackage{booktabs}       
\usepackage{amsfonts}       
\usepackage{nicefrac}       
\usepackage{microtype}      
\usepackage{xcolor}         

\usepackage{wrapfig}
\usepackage{graphicx}
\usepackage{subcaption}
\usepackage{booktabs} 
\usepackage{multirow}
\usepackage{bm}
\usepackage{hyperref}

\usepackage{amsmath}
\usepackage{amssymb}
\usepackage{mathtools}
\usepackage{amsthm}

\usepackage[capitalize,noabbrev]{cleveref}

\theoremstyle{plain}

\theoremstyle{definition}

\theoremstyle{remark}

\usepackage[textsize=tiny]{todonotes}

\title{$SE(3)$ Equivariant Ray Embeddings for \\ Implicit Multi-View Depth Estimation}

\author{
  Yinshuang Xu\\
  University of Pennsylvania\\
  \texttt{xuyin@seas.upenn.edu} \\
  \And
  Dian Chen\\
  Toyota Research Institute\\
  \texttt{dian.chen@tri.global} \\
  \And
  Katherine Liu\\
  Toyota Research Institute\\
  \texttt{katherine.liu@tri.global} \\
  \AND 
  Sergey Zakharov \\
  Toyota Research Institute \\
  \texttt{sergey.zakharov@tri.global} \\
  \And
  Rares Ambrus \\
  Toyota Research Institute \\
  \texttt{rares.ambrus@tri.global} \\
  \And
  Kostas Daniilidis\\
  University of Pennsylvania\\
  \texttt{kostas@cis.upenn.edu}\\
  \and
  \textbf{Vitor Guizilini}\\
  Toyota Research Institute \\
  \texttt{vitor.guizilini@tri.global} \\
}

\begin{document}

\maketitle

\begin{abstract}
 Incorporating inductive bias by embedding geometric entities (such as rays) as input has proven successful in multi-view learning. However, the methods adopting this technique typically lack equivariance, which is crucial for effective 3D learning. Equivariance serves as a valuable inductive prior, aiding in the generation of robust multi-view features for 3D scene understanding. In this paper, we explore the application of equivariant multi-view learning to depth estimation, not only recognizing its significance for computer vision and robotics but also addressing the limitations of previous research. Most prior studies have either overlooked equivariance in this setting or achieved only approximate equivariance through data augmentation, which often leads to inconsistencies across different reference frames.
To address this issue, we propose to embed $SE(3)$ equivariance into the Perceiver IO architecture. 
We employ Spherical Harmonics for positional encoding to ensure 3D rotation equivariance, and develop a specialized equivariant encoder and decoder within the Perceiver IO architecture.
To validate our model, we applied it to the task of stereo depth estimation, achieving state of the art results on real-world datasets without explicit geometric constraints or extensive data augmentation.

\end{abstract}

\section{Introduction}
Equivariance is a valuable property in computer vision, leveraging various symmetries to reduce sample and model complexity while boosting generalization. It has seen broad application in fields such as 3D shape analysis \cite{thomas2018tensor, weiler20183d, deng2021vector}, panoramic image prediction \cite{cohen2018spherical,xu2022unified,esteves2023scaling}, and robotics \cite{simeonov2022neural, huang2023edge, ryu2022equivariant,brehmer2023edgi}. In particular, there is an increasing interest in equivariant scene representation from multiple viewpoints \cite{safin2023repast,xu2023se}, as the multi-view setting is a fundamental challenge in the field and equivariant representations are desirable for their robustness and efficiency. 

Meanwhile, multi-view depth estimation has always been a core topic in computer vision. Previous works \cite{im2019dpsnet, kusupati2020normal, huang2018deepmvs} leverage the explicit geometric constraint to construct the feature cost-volume for depth prediction. Recently, the paradigm of combining implicit representations with generalist architectures has been widely adopted and gaining success. Inserting inductive bias via the embedding of geometric entities (rays) in the multi-view setting \cite{zhang2024cameras,suhail2022generalizable,sajjadi2022scene} has become popular. Notably, in multi-view depth estimation, \citet{yifan2022input} effectively combined geometric epipolar embeddings with image features for stereo depth estimation, outperforming traditional methods that depend on explicit geometric constraints. State-of-the-art work by \cite{guizilini2022depth} integrated multi-view geometric embeddings with image features for video depth estimation.  These methods show that the implicit multi-view geometry learned by the Perceiver IO architecture, which is a more efficient general architecture compared to the vision transformer~\cite{dosovitskiy2020image}, 
can improve upon approaches that rely on traditional explicit geometric constraints, such as cost volumes, bundle adjustment, and projective geometry. However, the implicit multi-view geometry promoted by the Perceiver IO architecture lacks equivariance.  This limitation becomes apparent when transforming the coordinate frame representing input geometry, such as camera poses, ray directions, or 3D coordinates. These transformations change the input in such a way that non-equivariant architectures are unable to achieve the same results, as shown in Figure \ref{motivation}. 
 \input{figures/teaser}

 Although \cite{guizilini2022depth} have tried to approximate equivariance through extensive data augmentation, achieving true equivariance at an architectural level remains an ongoing challenge.  In this paper, we propose to embed equivariance with respect to $SE(3)$ transformation of the global coordinate frame, i.e., gauge equivariance, to the Perceiver IO model.  
We substitute traditional Fourier positional encodings for the ray embedding  with Spherical Harmonics, which are more suitable to represent 3D rotations. We custom-develop a $SE(3)$ equivariant attention module to seamlessly interact with different types of equivariant features. This is achieved using a combination of invariant attention weights and equivariant fundamental layers. During the decoding stage, this equivariant latent space is disentangled into the equivariant frame and invariant global features. Our approach not only simplifies the integration of existing modules without requiring a specialized design, but also allows the network to focus on effective scene analysis via an invariant latent space, reducing the effects of global transformations.
%
The equivariant frame is used to ``standardize'' the query ray, serving as an invariant input for the decoder. This method ensures that both sets of inputs for the decoder are invariant, leading to an invariant output regardless of the decoder used. Consequently, we can employ the conventional Perceiver IO decoder in our equivariant framework.
In summary, our key contributions are as follows:
\begin{itemize}
\item We integrate $SE(3)$ equivariance into a multi-view depth estimation model by design,  using spherical harmonics as positional encodings for ray embeddings, as well as a specialized equivariant encoder.
\item By leveraging the equivariant learned latent space, we introduce a novel scene representation scheme for multi-view settings, featuring a disentangled equivariance frame and an invariant scene representation.
\item We assess our model's ability to learn 3D structures through wide-baseline stereo depth estimation. Our model delivers state-of-the-art results in this task, significantly surpassing the non-equivariant baseline.
\end{itemize}
\input{figures/diagram}

\section{Related Work}

\paragraph{Equivariant Networks} Equivariant Networks are garnering interest in vision for their efficiency and powerful inductive bias. These networks can be categorized by the data structures over which they operate, spanning 2D images~\cite{esteves2017polar,rahman2023truly}, graphs~\cite{satorras2021n,puny2023equivariant}, 3D point clouds~\cite{zhu2023e2pn, chen2021equivariant}, manifolds~\cite{cohen2019gauge,pim2021gauge}, and spherical images~\cite{esteves2020spin,cobb2020efficient}. From an architectural perspective, methods can also be classified by the tools they rely on, such as group convolution~\cite{cohen2016group,esteves2019equivariant,macdonald2022enabling,finzi2020generalizing}, steerable convolution on homogeneous spaces~\cite{worrall2017harmonic,weiler2019general,weiler20183d,cohen2019general,xu2022unified}, and recently transformers~\cite{romero2020group,romero2020attentive,liao2022equiformer,he2021gauge}. In the context of this paper, we highlight significant $SE(3)$ equivariant transformer works. \citet{fuchs2020se} first introduced an SE(3) equivariant transformer for point clouds, using steerable kernels for transformers and focusing on local features in point clouds. 
\citet{liao2022equiformer,liao2023equiformerv2} adopted a message-passing architecture for 3D equivariant transformers in point clouds. 
\citet{xu2023se} applied similar techniques for ray space. 
Our approach differs by using direct input-level positional encodings, rather than modifying the kernel with relative poses. 
We learn a global, non-hierarchical representation. 
\citet{safin2023repast} inserts pairwise relative poses in self-attention with a conventional attention module, requiring quadratic computation and lacking compact scene representation, unlike our method. 
Closely related to our work, \citet{assaad2022vn} proposed vector neuron transformers embedded in the Perceiver IO encoder for point clouds. 
However, they replace the original latent array with a learnable transformation, did not use spherical harmonics for equivariant positional encoding,  or design an equivariant decoder within the Perceiver IO framework. %
We treat the original latent array as invariant, and learn a disentangled representation for the decoder with versatile queries.
%
\citet{esteves2021generalized} uses spherical harmonics for positional encoding, primarily to enhance spherical function learning, not for equivariance.
\vspace{-3mm}
\paragraph{Implicit Multi-View Geometry} Even in the age of deep learning, traditional multi-view stereo methods like COLMAP \cite{fisher2021colmap} are still widely used for structure-from-motion. These methods are accurate but slow due to complex post-processing steps. To speed things up while maintaining accuracy, learning-based methods adapt traditional cost volume-based techniques for depth estimation \cite{kendall2017end, carreira2022hip,huang2018deepmvs,im2019dpsnet}.  
Recently, transformers \cite{vaswani2017attention} have become prevalent approaches, replacing CNNs in terms of popularity and performance. The Stereo Transformer \cite{li2021revisiting} replaces cost volumes with an attention-based matching procedure inspired by sequence modeling.
IIB \cite{yifan2022input} leverages Perceiver IO \cite{jaegle2021perceiver} for generalized stereo estimation by incorporating the epipolar geometric bias into the model. \citet{liu2022petr, chen2023viewpoint} inject 3D geometry into the transformer akin to IIB for object detection, while \citet{chen2023viewpoint} learns equivariance in a data-driven way. A closely related study to ours is DeFiNe \cite{guizilini2022depth}, in which camera information is incorporated into Perceiver IO and used to decode predictions from arbitrary viewpoints. 
However, their approach relies on data augmentation to approximate equivariance in the Perceiver IO, whereas our design inherently ensures equivariance at an architectural level. 

\section{Method}

In this section we start with some preliminaries about Perceiver IO and our baseline, Depth Field Networks (DeFiNe) \cite{guizilini2022depth}, a state-of-the-art method integrating camera geometries into Perceiver IO for multi-view depth estimation. 
We then outline the concept of equivariance in multi-view scenarios in Section \ref{equi_def}. Given these preliminaries, in Section \ref{equi_posenc} we delve into the details of our proposed equivariant positional encoding for rays, in Section \ref{equi_attention} we elaborate on the attention mechanisms used in our model, and in Section \ref{equi_encoder} we describe our choice of encoder parameterization. Finally, in Section \ref{decoder} we describe our decoder procedure, focusing on the task of depth estimation. The pipeline of our proposed $SE(3)$ equivariant model in multi-view context is shown in Figure \ref{pipeline}. 

\subsection{Preliminaries: Input-level Inductive Biases to Perceiver IO}
The Perceiver IO \cite{jaegle2021perceiver} is a generalist transformer architecture that encodes input data $\mathcal{I}\in \mathbb{R}^{N_i\times C_i}$ into a latent space $\mathcal{R}\in \mathbb{R}^{N_R\times C_R}$ by cross-attending $\mathcal{I}$ with $\mathcal{R}$. Further refinement of this latent space $\mathcal{R}$ is achieved using self-attention layers, followed by cross-attention to decode predictions $\mathcal{O}\in \mathbb{R}^{N_o\times C_o}$ using queries $\mathcal{Q}\in \mathbb{R}^{N_o\times C_q}$. 
Many works exploit its generic nature by introducing inductive biases at an input level, namely, providing prior knowledge about the data for implicit reasoning. Specifically, DeFiNe \cite{guizilini2022depth} uses camera geometries to construct 3D positional encodings for the multi-view problem. Given $N$ images $\{I_i\}_{i=1}^{N}$ from a set of cameras with poses $\{T_i\}_{i=1}^{N}$ and intrinsics $\{K_i\}_{i=1}^{N}$, DeFiNe calculates 3D rays $\{r_{uv}^i\}_{uv=(1,1)}^{(H,W)}$ from each camera center $t_i$ to each pixel $(u, v)$ on image $I_{i}$, and obtains positional encodings $PE(r_{uv}^i, t_i)$ with a mapping $PE(\cdot)$. 
These positional encodings are combined with image embeddings $\mathcal{F} =\{f^i_{uv}\}$ from a visual feature extractor to be encoded by $\mathcal{R}$ such that:
\begin{align*}
\tiny
    &\mathcal{R}_1 = \textrm{cross-attn}(\mathcal{R}_0, \{f_{uv}^i \oplus PE(r^i_{uv},t_i)\}) \\
   & \mathcal{R}_k = \textrm{self-attn}(\mathcal{R}_{k-1}), \quad k=2,\dots,K
\end{align*}
To obtain predictions for a set of $M$ novel viewpoints, we can similarly calculate 3D query rays from poses $\{T'_j\}_{j=1}^{M}$ and intrinsics $\{K_j\}_{j=1}^{M}$ and map them to query positional encodings $\mathcal{Q}=\{PE(r_{uv}^j, t_j)\}$, which will be used to decode the latent space $\mathcal{R}$ via cross attention: $\mathcal{O} = \textrm{cross-attn}(\mathcal{Q}, \mathcal{R}_K)$. 
In this way, prior knowledge, i.e., 3D camera geometries, is directly fed into the model as additional input features for the implicit learning of multi-view geometry.

\subsection{Equivariance Definition in Multiview Context}
\label{equi_def}
After introducing the input-level inductive bias framework, it is worth noting that the poses of the encoding cameras, as well as the query viewpoints, are defined in a global reference frame $T_{\textrm{G}}$. However, this choice of global reference frame is subject to change, and the property of equivariance ensures that predictions remain identical under these changes. Assuming the global reference frame undergoes a transform $T^{-1} \in SE(3)$ to $T_{\textrm{G}}'=T^{-1}T_{\textrm{G}}$, the ray representations become $(Rr^{i(j)}_{uv},Rt_{i(j)}+t)$ when representing $T=(R,t)$. The equivariant model $\Phi$ should satisfy  
\begin{align*}
&\Phi(\{f^i_{uv}\oplus PE(Rr^i_{uv},Rt_i+t)\}, \{PE(Rr^j_{uv},Rt_j+t)\})\\
&= \Phi(\{f^i_{uv}\oplus PE(r^i_{uv},t_i)\}, \{PE(r^j_{uv},t_j)\}).
\end{align*}
For further details on the definition of the equivariance, please see Appendix \ref{App:equi_def}.

\subsection {Equivariant Positional Encoding}
\label{equi_posenc}
To ensure the positional encoding process is equivariant w.r.t a transformation group $G$, we would like to enforce that
$\Phi(\cdot, PE(\rho^{\mathcal{X}}_g x)) = \Phi(\cdot, \rho^{\mathcal{Y}}_g PE(x))$
for any $g \in G$, where $\rho^{\mathcal{X}}$ is the group representation on coordinate space,  and $\rho^{\mathcal{Y}}$ is the group representation on the positional encoding space. The traditional Fourier basis is translationally equivariant, as detailed in Appendix \ref{fourier}. \citet{kitaev2020reformer} used this to attain translational equivariance, employing a conjugate product for invariant attention. However, this approach lacks rotational equivariance.

\begin{figure}[t!]
\begin{center}
\centerline{\includegraphics[width=0.95\columnwidth]{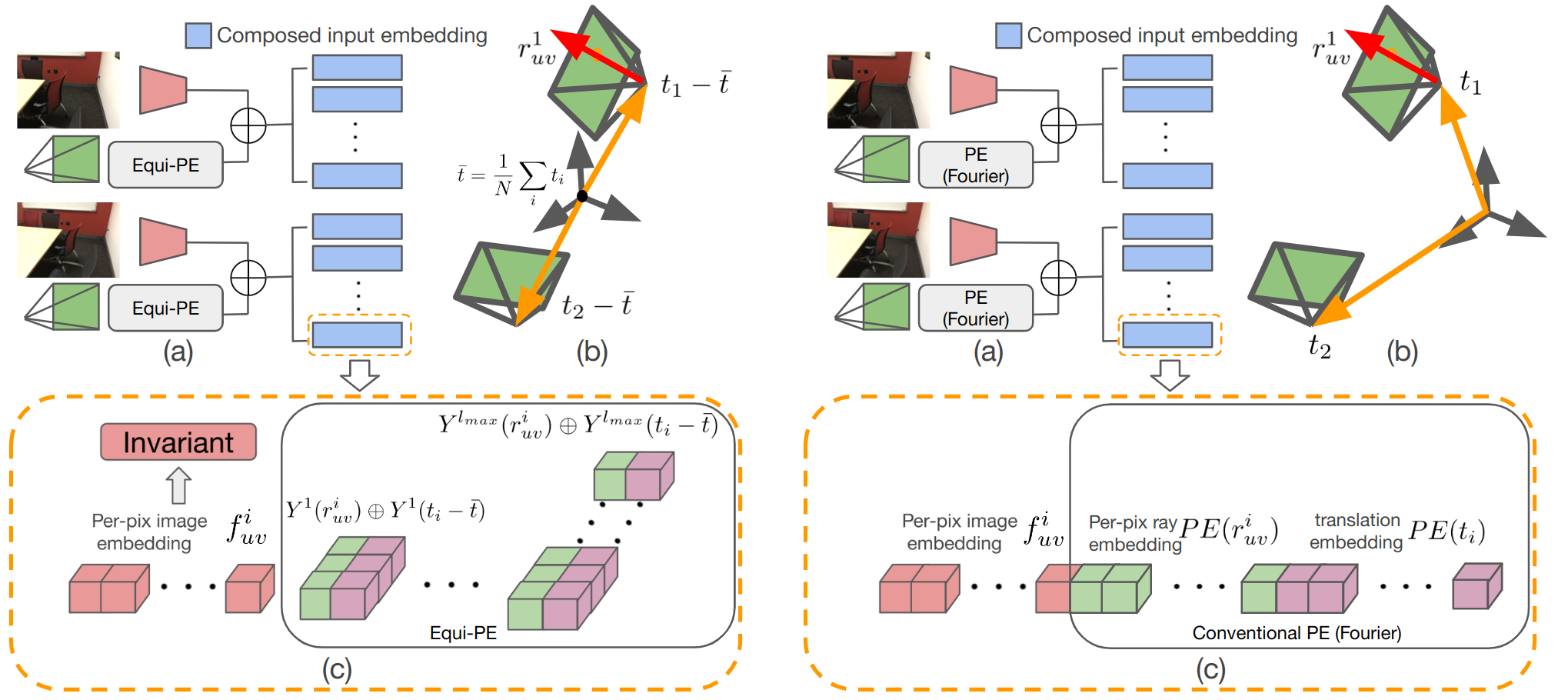}}
\caption{Comparison between an equivariant input embedding in our model (left) and the conventional input embedding in DeFiNe (right). (a) Pipeline used to generate input embeddings for the encoder, resulting in cross-attention keys and values. (b) To generate geometric information, we calculate embeddings for each ray $r^i_{uv}$ and camera relative position $t_i -\bar{t}$; (c) The final composed embedding format includes both image embeddings, which are invariant, and geometric embeddings, which are equivariant. In contrast, the conventional approach by Perceiver IO, as highlighted in parts (a) and (c), integrates Fourier positional encodings with image embeddings to form the input embeddings. Furthermore, as indicated in (b), Perceiver IO utilizes each ray $r^i_{uv}$ and the absolute translation $t_i$ for positional encoding purposes.}
\label{input_embedding}
\end{center}
\end{figure}

 This raises a key question: Are there any basis functions equivariant to both 3D translations and rotations? Unfortunately, none exist. However, a common method in equivariant works for translational equivariance is to subtract the center point, a technique we apply in our context as illustrated in Figure \ref{input_embedding}. This enables translational invariance, leaving the model to focus solely on achieving rotational equivariance. To address the 3D rotational equivariance, we turn to spherical harmonics (SPH), known for their inherent rotation-equivariant properties. They offer a way to accommodate 3D rotational changes, thereby achieving $SE(3)$ equivariance.
\subsubsection{Spherical Harmonics}
We provide a detailed introduction to Spherical Harmonics in Appendix \ref{spherical_harmonics}, where we also discuss how their application in previous equivariant transformers differs from their use in our model. Below, we present a brief overview of Spherical Harmonics for clarity. Similar to the varying frequencies of sines and cosines in Fourier series, spherical harmonics are characterized by different degrees (orders), denoted as $l \in \mathbb{N}$.  An order-$l$ spherical harmonics, denoted as $Y^l: \mathbb{R}^3 \rightarrow \mathbb{R}^{2l+1}$, 
follows the transformation rule: $Y^l(Rr)=D^l(R)Y^l(r),
Y^l(r)=\|r\|^lY^l(\hat{r})$,
where $R \in SO(3)$, $\hat{r}$ is the unit vector, 
$D^l: SO(3) \rightarrow \mathbb{R}^{(2l+1)\times(2l+1)}$ is called the Wigner-D matrix, serving as the irreducible representation of $SO(3)$ corresponding to the order $l$.  The Wigner-D matrix is an orthogonal matrix, that is $D^l(R)(D^l(R))^T=I$. These important properties allow us to achieve equivariance in the Perceiver IO transformer architecture. 

\subsubsection{Equivariant Hidden features}
\label{hidden_feature}
In our model, we embed both camera centers and viewing rays using spherical harmonics. The embedding is given by: $PE(r^i_{uv}, t_i) = \bigoplus_{l \in L} (Y^l(r^i_{uv}) \oplus Y^l(t_i - \bar{t}))$,
where each part corresponds to the same order of spherical harmonics (Figure \ref{input_embedding}). Here, $ L = \{1, 2, \ldots, l_{max}\}$ and $\bar{t} = \frac{1}{N} \sum_{i=1}^N t_i $, highlighting the extraction of the cameras' central position for translational invariance.
Due to the properties of spherical harmonics, 
the positional encoding of transformed input, 
$PE(Rr^i_{uv}, Rt_i+t)$, is equal to $\bigoplus_{l \in L} (D^l(R)(Y^l(r^i_{uv}) \oplus Y^l(t_i - \bar{t})))=R \cdot PE(r^i_{uv},t_i)$,
for any rotation $R \in SO(3)$ and translation $t \in \mathbb{R}^3$. In other words, it guarantees that these embeddings are both \emph{rotationally equivariant} and \emph{translationally invariant}. The transformation of these embeddings operates by multiplying each block with its respective Wigner-D matrix.

The image remains unchanged when the reference frame is transformed, as its contents are unaffected. Mathematically, this property is akin to multiplying by a $0$-order Wigner Matrix, equivalent to an identity. Thus, combining image features with positional encodings (Figure \ref{input_embedding}) transform as:
\vspace{-2mm}
\begin{align*}
\small
(f^i_{uv}, PE(Rr^i_{uv},Rt_i+t)) &= (D^{0}(R)f^i_{uv}, R \cdot PE(r^i_{uv}, t_i))\\
&= R \cdot (f^i_{uv},PE(r^i_{uv}, t_i))
\end{align*}
In our model, the equivariant hidden features mirror the structure of our embeddings, composed as $\bigoplus_{l \in L} H_l$
with subscripts indicating the feature type and $L=\left \{0,1,\cdots, l_{max}\right \}$. The size for each feature type $H_l$ follows $(2l+1, C_l)$, where $2l + 1$ is the intrinsic dimension and $C_l$ is the number of channels. For more a intuitive understanding, we visualize these embeddings in Appendix \ref{equi_latent} (Figure \ref{transformation_of_latent}).
Similar to the input embeddings, any rotation $R$ in $SO(3)$ rotates the hidden features as 

\begin{align*}
\small
R\cdot \bigoplus_{l \in L} H_l = \bigoplus_{l \in L} D^l(R) H_l
\end{align*}
where $D^l$ are the Wigner-D matrices. We disregard any translation action since the input and queries become translation-invariant after center subtraction.
\subsection{Basic Attention Modules}

\label{equi_attention}
This section highlights the equivariant attention module, fundamental to ensure encoder equivariance as depicted in Figure \ref{attention_module}. It consists of basic equivariant layers and an invariant multi-head inner product. Our architecture, unlike typical equivariant transformers \cite{fuchs2020se,liao2022equiformer}, does not enforce geometric constraints in the equivariant kernel. Instead, it incorporates all geometric features at an input-level. Our module, utilizing the Perceiver IO structure, learns global latent representations, in contrast to other methods that emphasize the hierarchical learning of local features. 

\paragraph{Equivariant Foundamental Layers} For the fundamental layers, we use the equivariant linear layer and layer normalization commonly used in previous works \cite{thomas2018tensor, fuchs2020se,liao2022equiformer,liao2023equiformerv2}, and provide additional details in Appendix \ref{app_fundamental}.  For equivariant nonlinear layers, there have been multiple proposed methods for features with the same format as ours: Norm-based Nonlinearity, Gate Nonlinearity, and Fourier-based Nonlinearity. Here, we take inspiration from the nonlinearity of Vector Neuron \cite{deng2021vector} and adapt a similar vector operation to higher-order features. Please see Appendix \ref{nonlinearity} for details of equivariant nonliearity. 
To better understand the differences between the basic layers in equivariant attention module and those in conventional one, we have visualize them in Appendix \ref{app_fundamental} and Appendix \ref{nonlinearity}. 

\paragraph{Multi-Head Attention Inner Product} As done in previous equivariant transformer works \cite{fuchs2020se,liao2022equiformer}, we can obtain the invariant attention matrix through inner product of the same types of features. These transformers that emphasize the hierarchical learning of local features suggest using tensor products of edge feature and node feature to mix different feature types, which is computationally demanding. We discard the tensor product and only calculate attention weights using various feature types and then multiply these weights with multi-type features to efficiently integrate different types of feature.  Please see Appendix \ref{App:MH_attention_inner} for more details.
Alternatively, we can mix feature types by treating them as Fourier coefficients for spheres, apply transformers on the sphere, and use Fourier Transform to obtain new coefficients. Please refer to Appendix \ref{alter_attn} for details.

\input{figures/mh_and_equi_array}

\subsection{Equivariant Encoder}
\label{equi_encoder}
\subsubsection{Equivariant Cross-attention}
As shown in the left part in Figure \ref{input_embedding} and Section \ref{hidden_feature}, the cross-attention input is in the format $\bigoplus_{l \in L} H_l$ with $C_l =2$ for $l\geq 1$ and $C_0$ being the channel number of $f^i_{uv}$. To facilitate a clearer understanding, a comparison of this input embedding with the one used in DeFiNE is also depicted in Figure \ref{input_embedding}. The latent array $\mathcal{R}_0 =((R_0)_1, (R_0)_2, \cdots, (R_0)_{N_R}) \in \mathbb{R}^{N_R \times C_R}$ can be treated as a scalar ($0$-order) feature, remaining constant during transformations in the reference frame.  To make the latent array also learn the geometric information, we apply a technique similar to \cite{assaad2022vn}, learning equivariant features from the input's averaged geometric information.  Specifically, we apply the positional encoding (PE) for each camera rotation, with each order being the concatenation of embeddings of the rotation matrix's three columns. The PE is then averaged over cameras. For the specific formulation please see Appendix \ref{average_eb}.

We obtain a global geometric latent $\mathcal{G}$ using an equivariant linear layer, where the size of the weight matrix $W_l$ for each type $l$ is $(3, N_RC^l)$, with $C^l$ being the channel count for type-$l$ feature in each latent. We then append this equivariant feature to the latent $\mathcal{R}_0$, forming a new latent array $\mathcal{R}'_0 = ((R'_0)_1, (R'_0)_2, \cdots, (R'_0)_{N_R})$, 
where $(R'_0)_i=(R_0)_i \oplus \bigoplus_{l\in L} (G_l)_i$ with $L=\left \{1,2,\cdots l_{max}\right \}$
and $(G_l)_i \in \mathbb{R}^{(2l+1)\times C^l}$. Figure \ref{latent_array} illustrates the construction of this equivariant latent array. With both an equivariant input embedding and latent array, we apply equivariant cross-attention to get the equivariant latent output.
\subsubsection{Equivariant Self-Attention}
\begin{wrapfigure}{r}{0.7\linewidth}
\includegraphics[width=0.4\textheight]{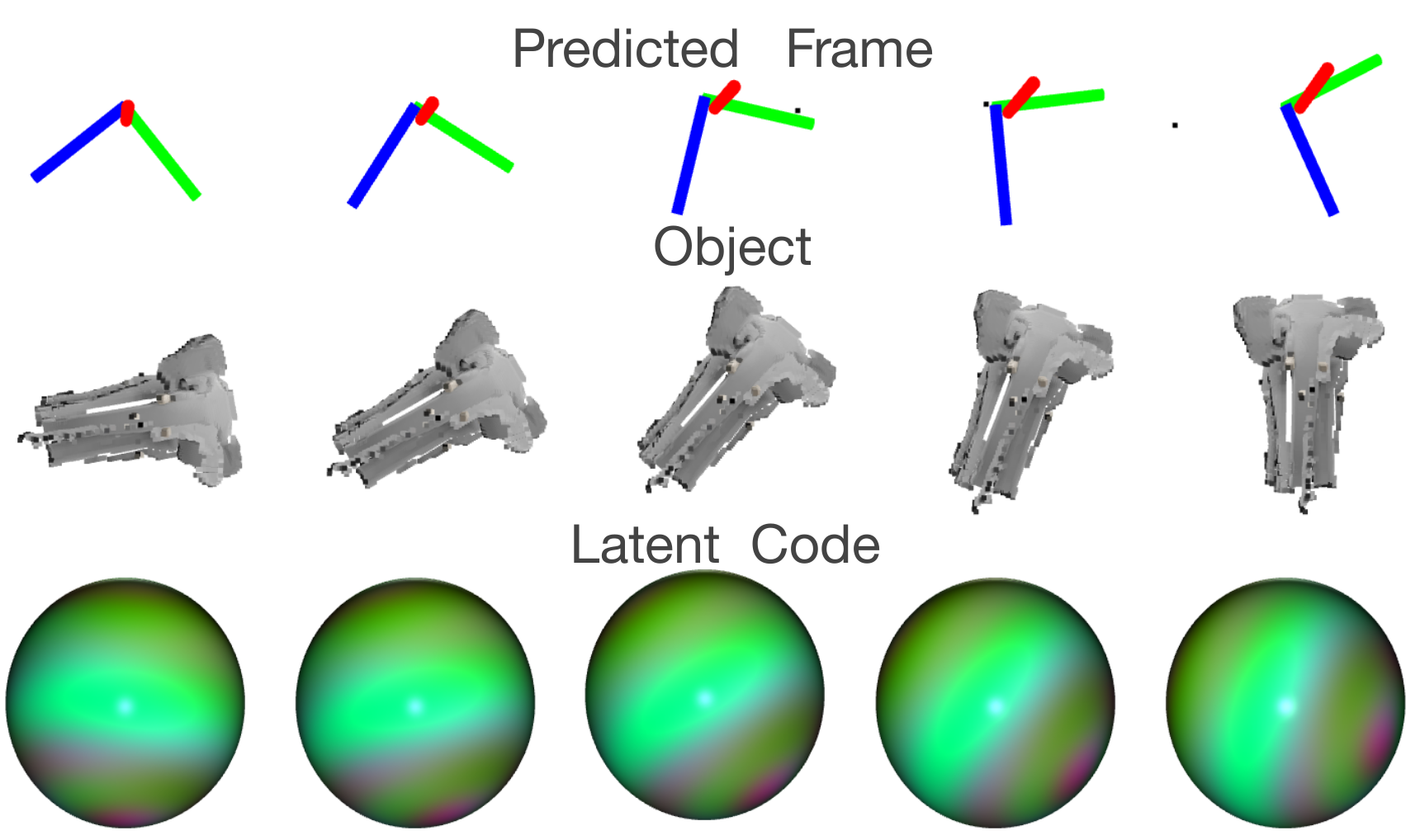}
         \caption{Equivariant latent code and predicted frame. For simplicity, we use object rotation to denote the inverse rotation of the reference frame. When the object is rotated, our latent code and predicted canonical frame are also rotated.}
   \label{equivariant_latent}
\end{wrapfigure}

We apply a self-attention equivariant attention mechanism to the equivariant output of cross-attention, producing a conditioned equivariant latent code. For visualization purposes (Figure \ref{equivariant_latent}), we can treat the equivariant latent code as the Fourier coefficients of spherical functions. Note that we do not map the features to a 2D color image. Since we have features with type-0, type-1, and type-2, etc, but we randomly select each channel from different types of feature and apply the inverse Fourier transform to get a spherical function and visualize it on a 3D sphere. For a proof of this result (i.e., the visualized sphere is also rotated when the latent code is rotated), 
please see Appendix \ref{visualization_sphere}.

\subsection{Decoder}
\label{decoder}
In Figure \ref{pipeline} we show that, before inputting the equivariant latent space and geometric query to the decoder, they are converted into an invariant form by establishing an equivariant frame. Specifically, the equivariant latent space $\mathcal{R}_K$ is represented as $\bigoplus_{l \in L} (\mathcal{R}_K)_l$. From its type-$1$ feature $(\mathcal{R}_K)_1$, we employ an equivariant MLP and the Gram-Schmidt orthogonalization~\cite{zhou2019continuity} to derive an equivariant frame, represented by a rotation matrix $R$. As depicted in Figure \ref{equivariant_latent}, the equivariant frame's rotation aligns with that of both the equivariant latent and the $3D$ scene.  Applying the inverse of $R$ to $\mathcal{R}_K$ results in a rotation-invariant latent code 
$\bigoplus_{l \in L} D^l(R)^T (\mathcal{R}_K)_l$, obtaining an invariant representation. See Appendix \ref{invariant_latent_query} for a proof.

For the embedding of camera center and viewing rays, we apply $R^T$ to $r^j_{uv}$ and $t_j - \bar{t}$ to obtain invariant coordinates (see Appendix \ref{invariant_latent_query} for a proof), denoted as $R^T r^j_{uv}$ and $R^T(t_j - \bar{t})$. We then use traditional sine and cosine positional encoding for these invariant coordinates, which allows us to leverage higher frequency information beyond the dimensional constraints of SPH. Since both the latent hidden state and the query are invariant to the transformation, this enables us to apply conventional cross-attention mechanisms to obtain invariant outputs and predictions, capturing higher frequency details and improving expressiveness. 
\section{Experimental Results}
\subsection{Datasets and Implementation}
\label{dataset}
We use \textbf{ScanNet}~\cite{dai2017scannet} and \textbf{DeMoN}~\cite{ummenhofer2017demon} to validate our model on the task of stereo depth estimation. For ScanNet, we use the same setting as \cite{kusupati2020normal}, which downsamples scenes by a factor of $20$ and splits them to obtain $94212$ training and $7517$ test pairs. The DeMoN dataset includes the SUN3D, RGBD-SLAM and Scenes11 datasets, where SUN3D and RGBD-SLAM are real world datasets and Scenes11 is a synthetic dataset. There are a total of $166285$ training image pairs from $50420$ scenes, and we use the same test split as \cite{kusupati2020normal} ($80$ pairs in SUN3D, $80$ pairs in RGBD and $168$ pairs in Scenes11). We include details on the network architecture and implementation in Appendix \ref{App:Network_Arch}.

\input{figures/depth}
\subsection{Stereo Depth Estimation}
We compare our equivariant model with other state-of-the-art methods on stereo depth estimation, and report quantitative results in Table \ref{quantitative result}. As we can see, it significantly outperforms competing methods on all real-world datasets and shows comparable results to the state-of-the-art on Scenes11, a synthetic dataset.
This superior performance on real-world datasets is evidence of the benefits of using equivariance in multi-view scene representation. Synthetic datasets, unaffected by real-world lighting conditions, camera miscalibration and view-dependent artifacts, might benefit approaches such as DPSNet \cite{im2019dpsnet} and NAS \cite{kusupati2020normal} that use cost volume to achieve view consistency. %
It's also noteworthy that NAS \cite{kusupati2020normal} uses additional ground truth surface normals as supervision.
\input{depth_eval_2}
\input{data_aug_new}
We denote our model with ``Equi'' and our baseline, DeFiNe, with ``Nonequi'' to highlight that both use the same architecture, Perceiver IO, with the key difference being the presence of equivariance in our model.

Additionally, to assess the advantages of incorporating equivariance into our model, we conducted a comparative analysis of our model against our nonequivariant DeFiNe baseline~\cite{guizilini2022depth}, both with and without data augmentation. 
We explore two kinds of data augmentation: virtual camera augmentation (VCA), in which novel viewpoints are generated via pointcloud reprojection; and canonical camera jittering (CCJ), in which the reference frame is perturbed with random rotation and translation, reported in Table \ref{compare_baseline}. 
To further showcase our equivariant properties, we visualize the predicted canonical frame and reconstructed $3D$ point clouds from depth maps. In Figure \ref{equivariance_reference}, we see that, for the same scene, when we switch the reference frame (in red) between cameras, the output point clouds change when using the standard Perceiver IO architecture, while ours remain constant, since the predicted depth is equivariant to transformations. Furthermore, even when we use different image pairs within the same scene, which theoretically cannot be guaranteed equivariant due to changes in image content, our model still predicts near-consistent canonical frames and point clouds, as illustrated in Figure \ref{approximate_equivariance}.In the meanwile, we compare our model with current state-of-art depth estimation model Depth anything \cite{yang2024depth}, and we provide the results in Appendix \ref{depth_anything_app}.  
\subsection{Ablation Study}
\label{ablation_study}
We performed an ablation study on the geometric positional encodings, spherical harmonics encoding, equivariant attention, and the decoder architecture, and report the quantitative results in Table \ref{tab:ablation}. As expected, when positional encoding is not used, results are significantly degraded due to missing geometric information. Our results demonstrate that the model leverages geometric information to learn implicit multi-view geometry.  
Although using Fourier positional encodings with our method breaks the equivariant properties, we conducted ablations by replacing spherical harmonics with Fourier encodings to assess the specific contribution of spherical harmonics.
\input{figures/pointcloud_new}

As shown in the table, Fourier encodings, which are not equivariant, are incompatible with an equivariant architecture, resulting in significantly worse performance. Additionally, we replaced the equivariant attention module with a conventional one in another ablation study.
Removing the equivariant attention layers also disrupts the equivariance of our architecture, leading to a substantial drop in performance since the model loses its theoretical equivariance. We also explore the impact of the maximum order of spherical harmonics in the positional encodings, indicated by $l_{max}$.
For network architecture, we evaluated the impact of not learning the canonical frame, that is, we use the equivariant attention module in the decoder followed by transferring the equivariant output to invariant output via inner product, see Appendix \ref{sec:alternative} for details.
We noticed that higher order of spherical harmonics improve depth estimation, since high frequency promotes fine-grained learning and differentiate positions in a higher-dimensional space. 
\input{ablation_new}
Unlike Fourier basis, the dimension of the spherical harmonics grows two times linearly with increasing order, which is a limitation of our method, and therefore we keep the highest SPH order as $8$ in our final model. 
This is also a reason why learning a equivariant canonical frame for invariant decoding with Fourier basis and a conventional decoder is a better approach than directly using an equivariant decoder. Another factor is that learning a canonical frame enforces all inputs to the decoder to be invariant, which should facilitate 3D reasoning. Moreover, we performed additional small-scale experiments to study the impact of the number of available views, see Appendix \ref{varying_view}.

\section{Conclusion and Discussion}
\label{App:limitation}
We introduce an $SE(3)$ equivariant model designed to learn the equivariant $3D$ scene prior across multiple views, utilizing spherical harmonics for positional encoding and specialized equivariant attention mechanisms within the Perceiver IO architecture.
Additionally, our design exploits its equivariant latent space to disentangle equivariant frames and invariant scene details, enabling the seamless integration of various existing decoders in conjunction with our specialized encoder.  our model's capability in 3D structure comprehension is showcased through its superior performance in stereo depth estimation, significantly exceeding that of non-equivariant models. Our architecture can be modified to accommodate a wider range of vision tasks, which we leave to future work (for a more detailed discussion please see Appendix \ref{general_task}). 

\paragraph{Limitation} As discussed in Section \ref{ablation_study}, unlike the Fourier basis, the dimension of spherical harmonics increases linearly at twice the rate with each order. This limits the number of spherical harmonics and the maximum frequency utilized, resulting in an inability to preserve detailed features in cameras and images. Additionally, the presence of different types of features, each with its own linear and nonlinear layers, slightly slows down the forward process compared to traditional methods. Moreover, we observe instability in training the equivariant network, which may be due to the magnitude explosion of high-order spherical harmonics. 

\newpage
\section*{Acknowledgment} 
This research was supported by Toyota Research Institute, whose funding and resources were invaluable in advancing this work. 

{\small
\bibliographystyle{plainnat}
\bibliography{example_paper}
}
\newpage
\appendix
\onecolumn
\section*{Appendix}
\input{appendix}

\newpage

\end{document}

%% file: figures/teaser.tex

\begin{wrapfigure}{r}{0.5\linewidth}
\includegraphics[width=0.3\textheight]{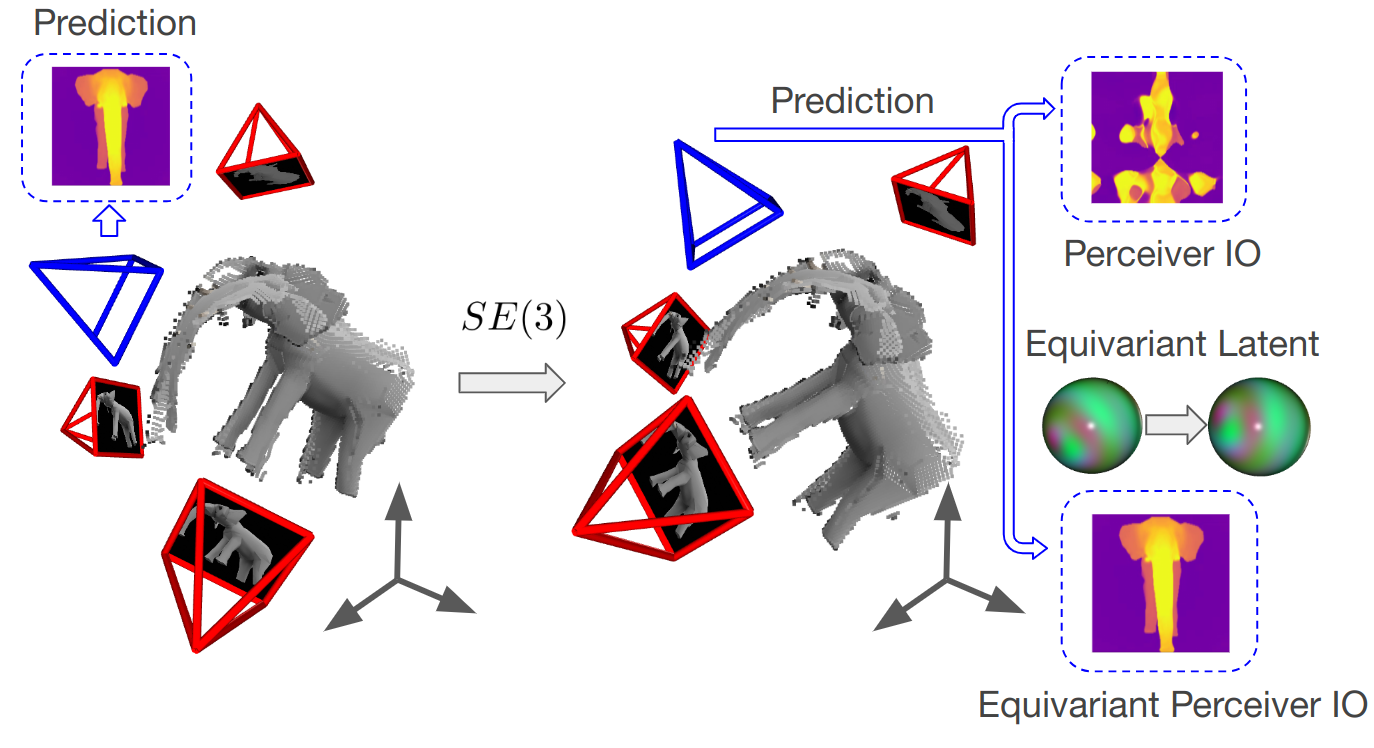}
         \caption{
         Given a sparse set of posed images (red), the task is to estimate depth for a novel viewpoint (blue). The Perceiver IO struggles to accurately predict depth when the reference frame (gray) changes, equivalent to an inverse transformation applied to the object and cameras. In contrast, our model delivers the consistent result due to its equivariant design.
         }
   \label{motivation}
   \vspace{-3mm}
\end{wrapfigure}

%% file: figures/diagram.tex
\begin{figure*}
\centering
\centerline{\includegraphics[width=0.99\textwidth]{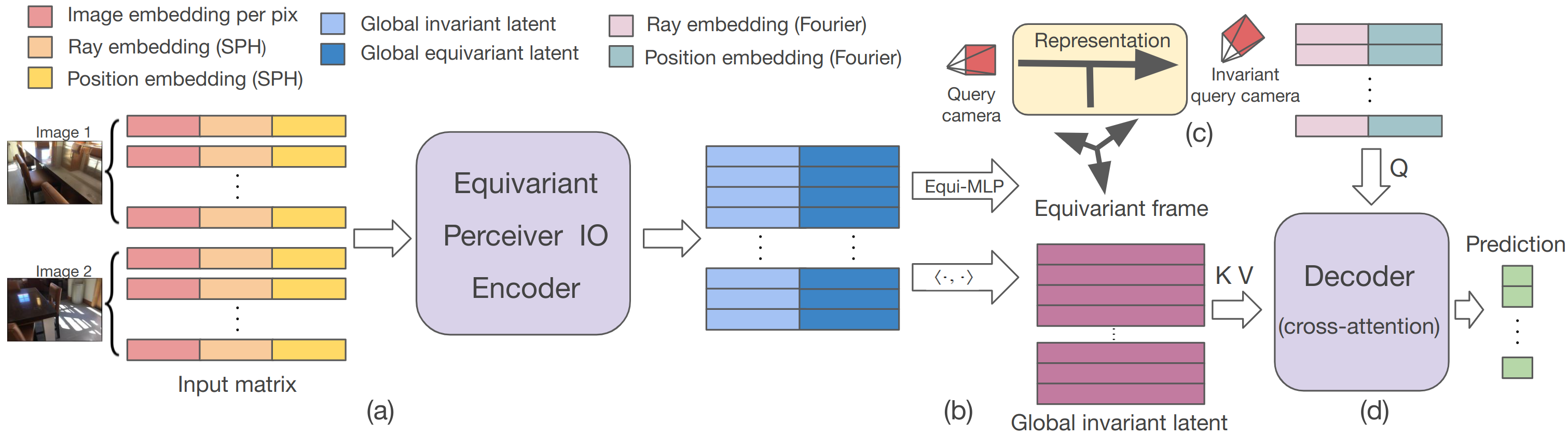}}
\caption{Our proposed Equivariant Perceiver IO (EPIO) architecture. (a) We take as input the concatenation of per-pixel image, ray, and camera embeddings, the latter two calculated using spherical harmonics. (b) The output of our equivariant encoder is a global latent code, including both global invariant and equivariant components. From those, we extract an equivariant reference frame through an equivariant MLP, while simultaneously obtaining invariant latents through inner product. (c) When a query camera is positioned in this equivariant reference frame, its pose becomes invariant, which enables the use of conventional Fourier basis to encode it. (d) Given an invariant latent and invariant pose, we use a conventional Perceiver IO decoder to generate predictions for each query ray.}
\label{pipeline}
\vspace{-5mm}
\end{figure*}

%% file: figures/mh_and_equi_array.tex
\begin{figure*}[t!]
    \centering
    \begin{subfigure}[b]{0.35\textwidth}
        \centering
        \includegraphics[height=1.7in]{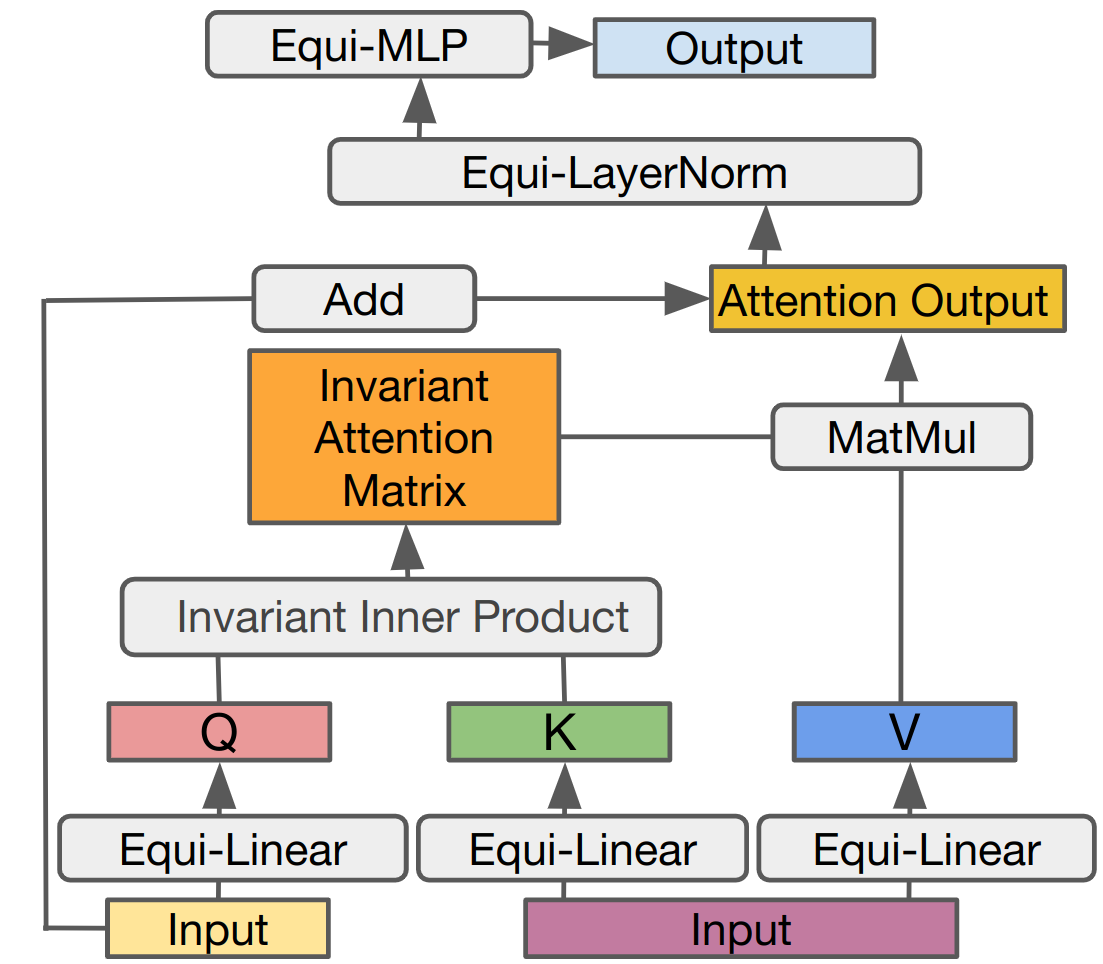}
        \caption{}
        \label{attention_module}
    \end{subfigure}
    \begin{subfigure}[b]{0.55\textwidth}
        \centering
        \includegraphics[height=1.4in]{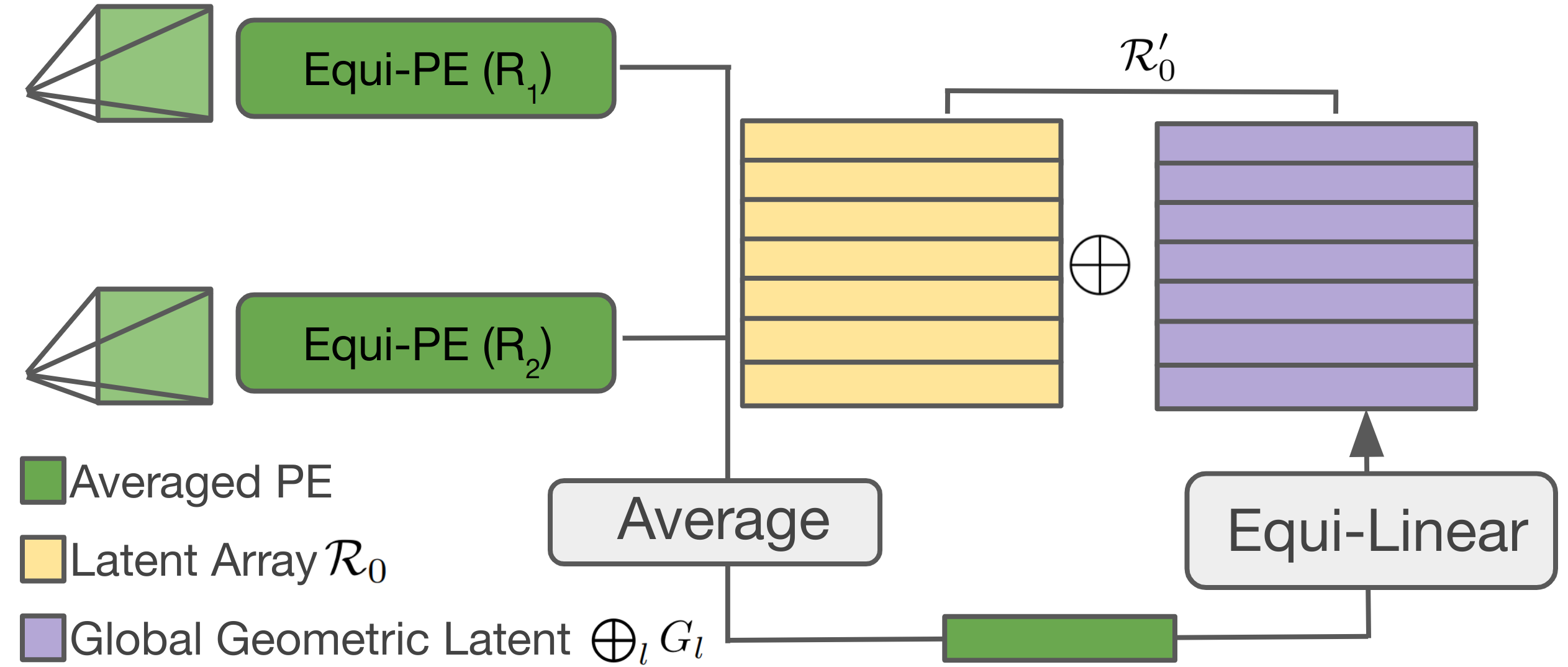}
        \caption{}
        \label{latent_array}
    \end{subfigure}
    \caption{Left: Our equivariant module is distinct from traditional implementations~\cite{vaswani2017attention} in its fundamental layers and the key-query product, that are crafted to be respectively equivariant and invariant. Right: Equivariant latent array used as additional input to the encoder. We apply equivariant positional encoding to each camera rotation, which is then averaged. We leverage an equivariant linear layer to get a global geometric latent $\oplus_lG_l$, which is concatenated with the conventional latent array $\mathcal{R}_0$ to compose our proposed equivariant latent array $\mathcal{R}'_0$.}
    \vspace{-4mm}
\end{figure*}

%% file: figures/depth.tex
\begin{figure*}
\centering
\centerline{\includegraphics[width=5.5in]{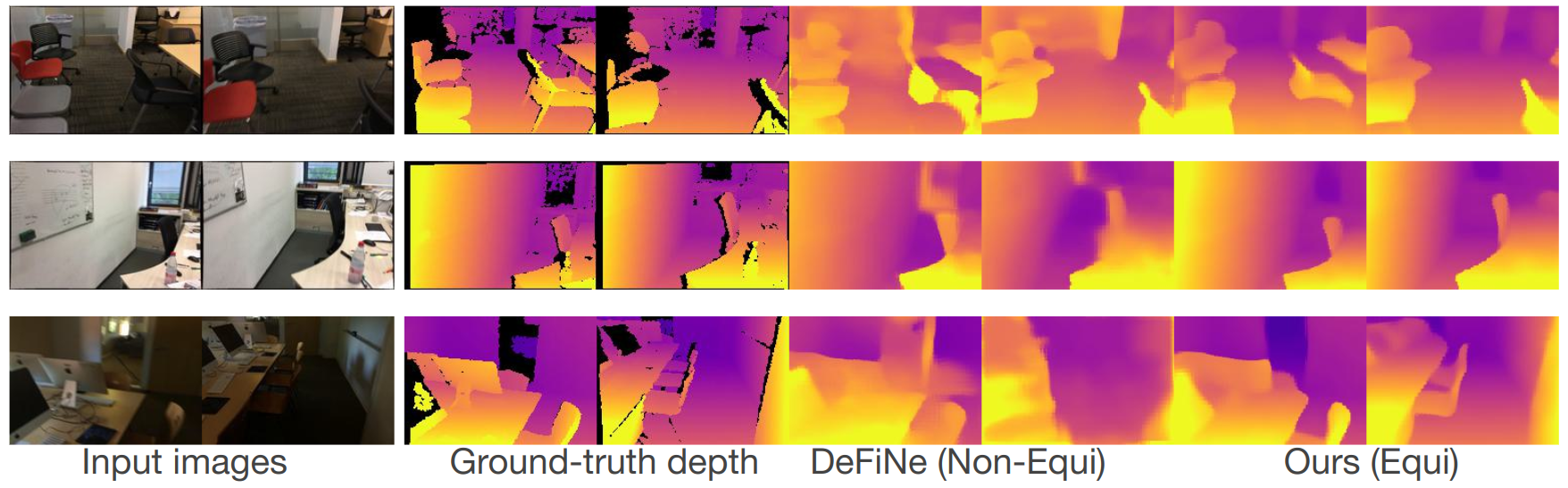}}
\caption{Stereo depth estimation results on ScanNet, using our proposed EPIO architecture.}
\label{qualitative results}
\end{figure*}

%% file: depth_eval_2.tex
\begin{table}[t]
\label{sample-table}
\begin{center}
\begin{small}
\setlength{\tabcolsep}{0.1em}
\renewcommand{\arraystretch}{0.95}
\begin{tabular}{l|cccc||l|ccccr}
\toprule
Dataset & Method & Abs.Rel. $\downarrow$ & RMSE $\downarrow$ & $\delta < 1.25\uparrow$& Dataset & Method & Abs.Rel. $\downarrow$ & RMSE $\downarrow$ & $\delta < 1.25\uparrow$\\
\midrule
\multirow{6}{*}{ScanNet}   & DPSNet & 0.126 & 0.315 &     - &\multirow{7}{*}{RGBD-SLAM} &   DeMoN & 0.157 & 1.780 & 0.801 \\
                           &    NAS & 0.107 & 0.281 &     - && DeepMVS & 0.294 & 0.868 & 0.549\\
                           &    IIB & 0.116 & 0.281 & 0.908 &&  DPSNet & 0.151 & 0.695 & 0.804\\
                           & DeFiNe & \underline{0.093} & \underline{0.246} & \underline{0.911}&&NAS & 0.131 & 0.619 & 0.857 \\
                           &  \textbf{Ours}  & \textbf{0.086} & \textbf{0.229} & \textbf{0.923}& &IIB & \underline{0.095} & \underline{0.550} & \underline{0.907} \\
                           &&&&&& \textbf{Ours} & \textbf{0.080} & \textbf{0.433} & \textbf{0.912} \\
\hline
\multirow{7}{*}{SUN3D }   &   DeMoN & 0.214 & 2.421 & 0.733 &\multirow{7}{*}{Scenes11}  &    DeMoN & 0.556 & 2.603 & 0.496 \\
                          & DeepMVS & 0.282 & 0.944 & 0.562 &&  DeepMVS & 0.210 & 0.891 & 0.688\\
                          &  DPSNet & 0.147 & 0.449 & 0.781 &&   DPSNet & \underline{0.050} & \underline{0.466} & 0.961 \\
                          &     NAS & 0.127 & 0.378 & 0.829 &&      NAS & \textbf{0.038} & \textbf{0.371} & \textbf{0.975}\\
                          &     IIB & \underline{0.099} & \underline{0.293} & \underline{0.902}&&      IIB & 0.055 & 0.523 & 0.963 \\
                          & \textbf{Ours} & \textbf{0.090} & \textbf{0.260} & \textbf{0.912}&& \textbf{Ours} & 0.069 & 0.617 & \underline{0.965}  \\
\bottomrule
\end{tabular}
\caption{Stereo depth estimation results compared with the state-of-the-art: DPSNet \cite{im2019dpsnet}, NAS \cite{kusupati2020normal}, IIB \cite{yifan2022input}, DeFine \cite{guizilini2022depth}, DeMoN \cite{ummenhofer2017demon}, DeepMVS \cite{huang2018deepmvs}.}
\label{quantitative result}
\end{small}
\end{center}
\end{table}

%% file: data_aug_new.tex
\begin{wraptable}{r}{0.7\textwidth}
\small
  \centering
\renewcommand{\arraystretch}{0.95}
\begin{tabular}{lcccr}
\toprule
Method & Abs.Rel. $\downarrow$ & RMSE $\downarrow$& $\delta < 1.25\uparrow$ \\
\midrule
DeFiNe (w/o VCA) 
& 0.117 & 0.291 & 0.870 \\
Ours (w/o VCA)
& 0.104 & 0.247 & 0.893 \\
\midrule
DeFiNe (w/o jitter)    
& 0.099 & 0.261 & 0.891 \\
DeFiNe     
& 0.093 & 0.246 & 0.911 \\
Ours (w/o jitter)
& \textbf{0.086} & \textbf{0.229} &  \textbf{0.923}     \\
\bottomrule
\end{tabular}
\caption{Comparison of our EPIO model and DeFiNe on ScanNet regarding the use of data augmentation. \emph{VCA} stands for \emph{virtual camera augmentation}, and \emph{jitter} stands for \emph{canonical camera jittering}.}
\label{compare_baseline}
\vspace{-5mm}
\end{wraptable}

%% file: figures/pointcloud_new.tex
    
    

\begin{figure*}[t!]
    \centering
    \begin{subfigure}[b]{0.43\textwidth}
        \centering
        \includegraphics[height=1.3in]{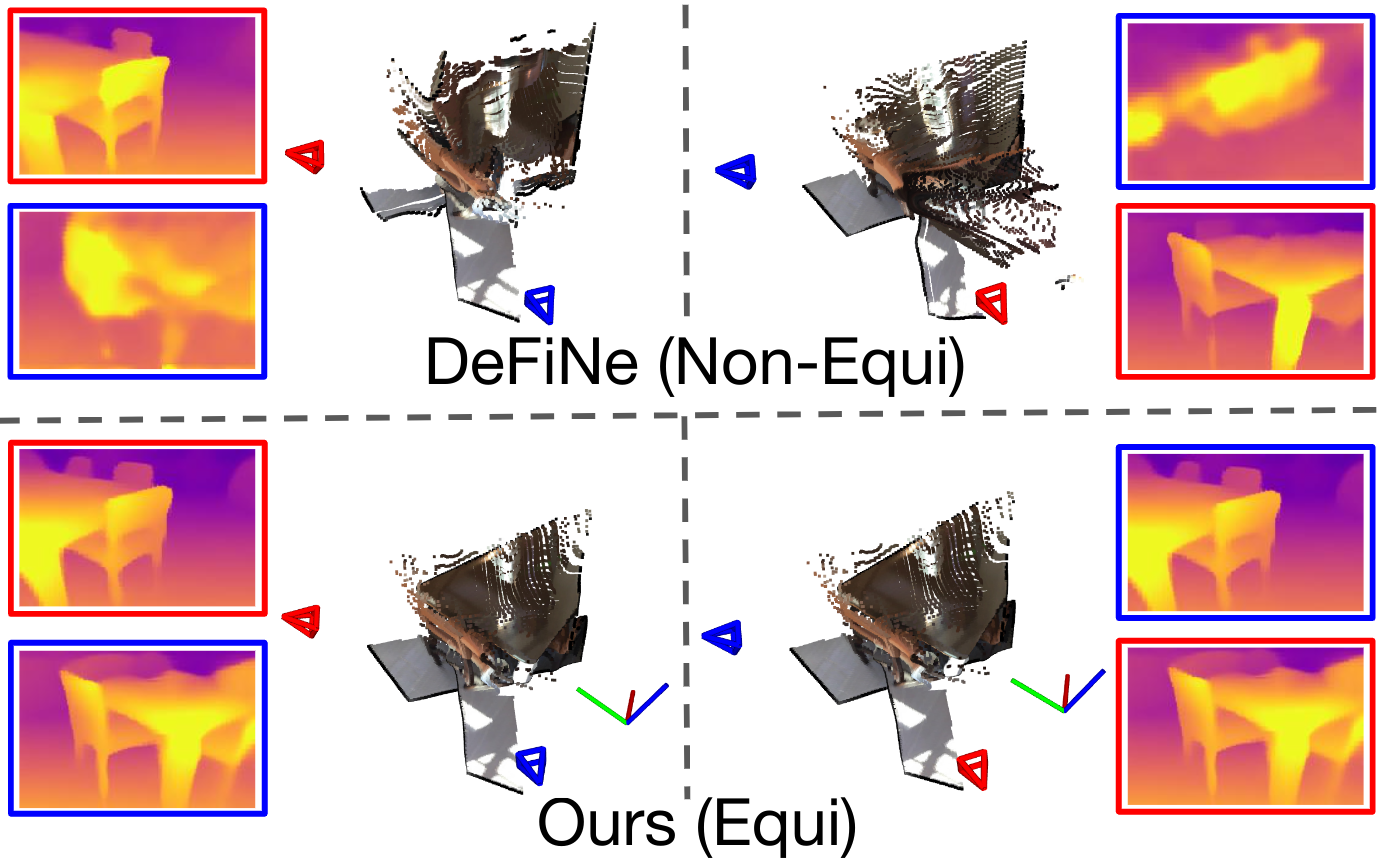}
        \caption{}
        \label{equivariance_reference}
    \end{subfigure}
    \begin{subfigure}[b]{0.5\textwidth}
        \centering
        \includegraphics[height=1.4in]{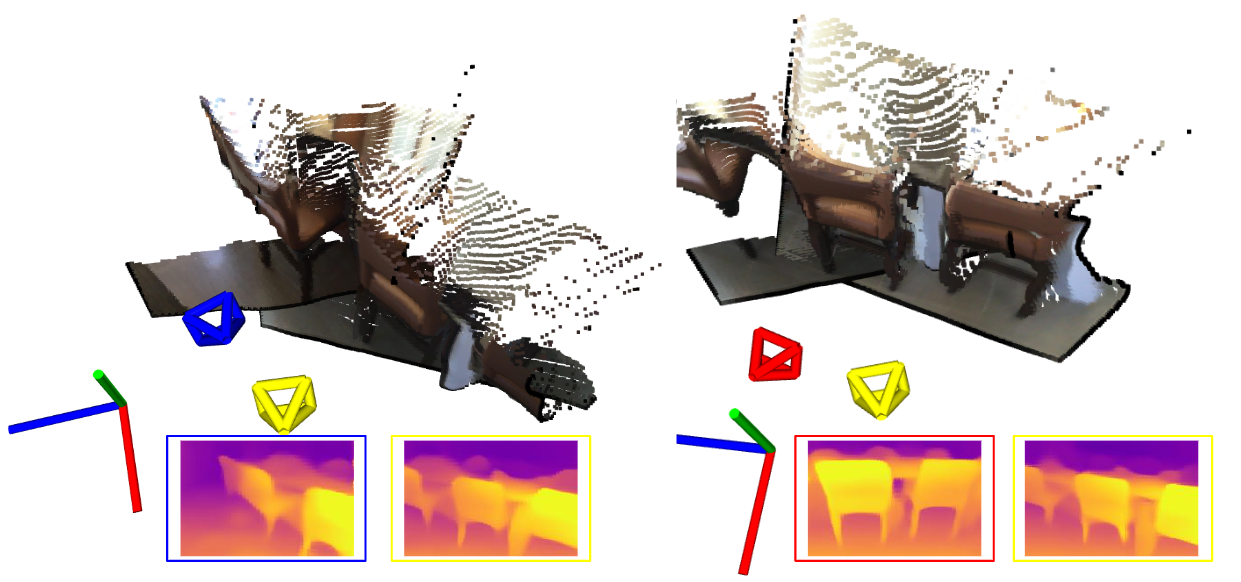}
        \caption{}
        \label{approximate_equivariance}
    \end{subfigure}
    \caption{The figure (a) shows the equivariance of changing reference frame (Red: reference frame): For the same input and varying camera frames as reference, the Perceiver IO's predictions change, but our model's predictions stay consistent and the predicted frame is equivariant to the reference frame transformation. The figure (b) shows the approximate equivariance for different camera sets.}
\end{figure*}

%% file: ablation_new.tex
\begin{wraptable}{r}{0.7\textwidth}
\tiny
  \centering
\renewcommand{\arraystretch}{0.95}
\small
\begin{tabular}{lcccr}
\toprule
Variation & Abs.Rel.$\downarrow$ & RMSE$\downarrow$ & $\delta < 1.25\uparrow$ \\
\midrule
w/o camera information  & 0.229&0.473& 0.661\\
w/ Fourier & 0.131&0.318& 0.843\\
w/o equi-attention & 0.127&0.314&0.851\\
Type $l_{max}=1$ &0.134&0.310 &0.869 \\
Type $l_{max}=2$   & 0.125&0.302& 0.875\\
Type $l_{max}=4$    & 0.116&0.283 &0.898\\
EquiDecoder    &0.128 &0.317 &0.857 \\
\midrule
\textbf{Full Model} &\textbf{0.086} &\textbf{0.229}&\textbf{0.923}\\
\bottomrule
\end{tabular}
\caption{Ablation study on the choice of positional encoding frequency and decoder architecture.}
\label{tab:ablation}
\vspace{-3mm}
\end{wraptable}

%% file: appendix.tex
\section{Preliminaries}
\subsection{Perceiver IO}
\label{app_perceiver}
The Perceiver IO \cite{jaegle2021perceiver} efficiently encodes multi-modality inputs by utilizing cross-attention between the inputs themselves and a learnable, fixed-dimension latent code. This latent code is then refined through a series of self-attention layers. In the decoding phase, the model employs cross-attention between a given query and the refined latent code to generate predictions.  

In DeFiNe \cite{guizilini2022depth}, this framework is used to address scenarios involving multiple cameras with predetermined relative poses, denoted as $\{T_i\}_{i=1}^N$, and their respective images $\{ I_i\}_{i=1}^N$. Within this context, the system queries arbitrary camera poses, represented as $\{T'_j\}_{j=1}^{M}$. Importantly, this queried pose $T_j$ can either be one of the already established camera positions (as it is common in stereo depth estimation), or a position outside the range of the input cameras (which is typical in video depth estimation).  Based on input data and queried pose, the network generates the corresponding predicted output $\hat{D}_j$ for the query.

Like other vision tasks that utilize the Perceiver framework, DeFiNe uses as input a composite of geometric information and corresponding image features, while the query utilizes only geometric information. Specifically, the input is formulated as $\left\{f^i_{uv} \oplus PE(r^i_{uv}) \oplus PE(t_{i})\right\}$, where $f^i_{uv}$ represents image features associated with each pixel 
$(u,v)$ in camera $i$. $PE(r^i_{uv})$ denotes the positional encoding with Fourier cosine and sine series of the ray direction $r^i_{uv}$ calculated relative to camera $T_i$, $PE(t_{i})$ refers to the positional encoding of the camera's translation $t_i$ in $T_i$, using Fourier cosine and sine series. The query in this model is represented as $\left\{PE(r^j_{uv}) \oplus PE(t_j)\right\}$, and the network is designed to output depth estimation $\hat{D^j_{uv}}$ for each query pixel $(u,v)$ of query camera $j$.

\subsection{Equivariance Definition}
\label{App:equi_def}
Concretely, assuming the global reference frame undergoes a transform $T^{-1} \in SE(3)$ to $T_{\textrm{G}}'=T^{-1}T_{\textrm{G}}$, the set of input and query poses would become $\{TT_i\}$ and $\{TT'_j\}$ with respect to $T_{\textrm{G}}'$ (note that the corresponding input images $\{I_i\}_{i=1}^{N}$ remain unchanged). Mathematically, we use $\Lambda_T((\{T_i\},\{I_i\})) = (\{TT_i\},\{I_i\})$  and $\Lambda_T(\{T'_j\}) = \{TT_j\}$ to denote the $SE(3)$ actions on the input and query cameras. An equivariant network satifies 
\begin{align*}
\small
\Phi(\Lambda_T(\{T_i\}, \{I_i\}), \Lambda_T(\{T'_j\})) \equiv \Phi( (\{T_i\}, \{I_i\}), \{ T'_j\}).
\end{align*}
Readers may recognize the above equation as describing the invariance of the network $\Phi$ to the transformation of both input and query. In fact, it is also equivalent to the statement that the network $\Phi$ is equivariant when viewed that it learns an implicit field. To demonstrate this equivalence, let $F(\cdot) = \Phi((\{T_i\}, \{I_i\}), \cdot)$, and define the $SE(3)$ operator $\Lambda'$ on $F$: $\Lambda'_TF(\{T_j\})= F(\Lambda_{T}^{-1}(\{T_j\}))$. We then derive  
\begin{align*}
\Phi(\Lambda_T(\{T_i\},\{I_i\}))=\Lambda'_TF = \Lambda'_T\Phi(\{T_i\},\{I_i\}),
\end{align*}
i.e., that the network $\Phi$ is equivariant. A similar statement can be found in \cite{chen20223d}.

With input level inductive bias and representing $T=(R,t)$, the equivariant model $\Phi$ should satisfy  
\begin{align*}
&\Phi(\{f^i_{uv}\oplus PE(Rr^i_{uv},Rt_i+t)\}, \{PE(Rr^j_{uv},Rt_j+t)\})\\
&= \Phi(\{f^i_{uv}\oplus PE(r^i_{uv},t_i)\}, \{PE(r^j_{uv},t_j)\}).
\end{align*}

\subsection{Fundamental Layers}
\label{app_fundamental}
\subsubsection{Equivariant Linear Layer}
With the transformation acting on the features, we can define the equivariant linear layer
$\mathcal{L}$:
\begin{align*}
\mathcal{L}(R \cdot H^{(k)}) = R \cdot \mathcal{L}(H^{(k)}) = R \cdot H^{(k+1)},
\end{align*}
where the superscript denotes the index of layer $H^{(k)} = \bigoplus_{l \in L} H^{k}_l =(H^{(k)}_0, H^{(k)}_1, \cdots, H^{(k)}_{l_{max}})$ and the same for $H^{(k+1)}$.
To achieve equivariance, we use the same linear layer $\mathcal{L}$ as stated in \cite{thomas2018tensor, weiler20183d, liao2022equiformer}:
\begin{align*}
&\mathcal{L}((H^{(k)}_0,H^{(k)}_1,\cdots, H^{(k)}_{l_{max}})) \\
&= (H^{(k)}_0 W_0, H^{(k)}_1W_1, H^{(k)}_{l_{max}}W_{l_{max}}).
\end{align*}

\begin{figure*}
\centering
\vspace{-10mm}
\centerline{\includegraphics[width=6in]{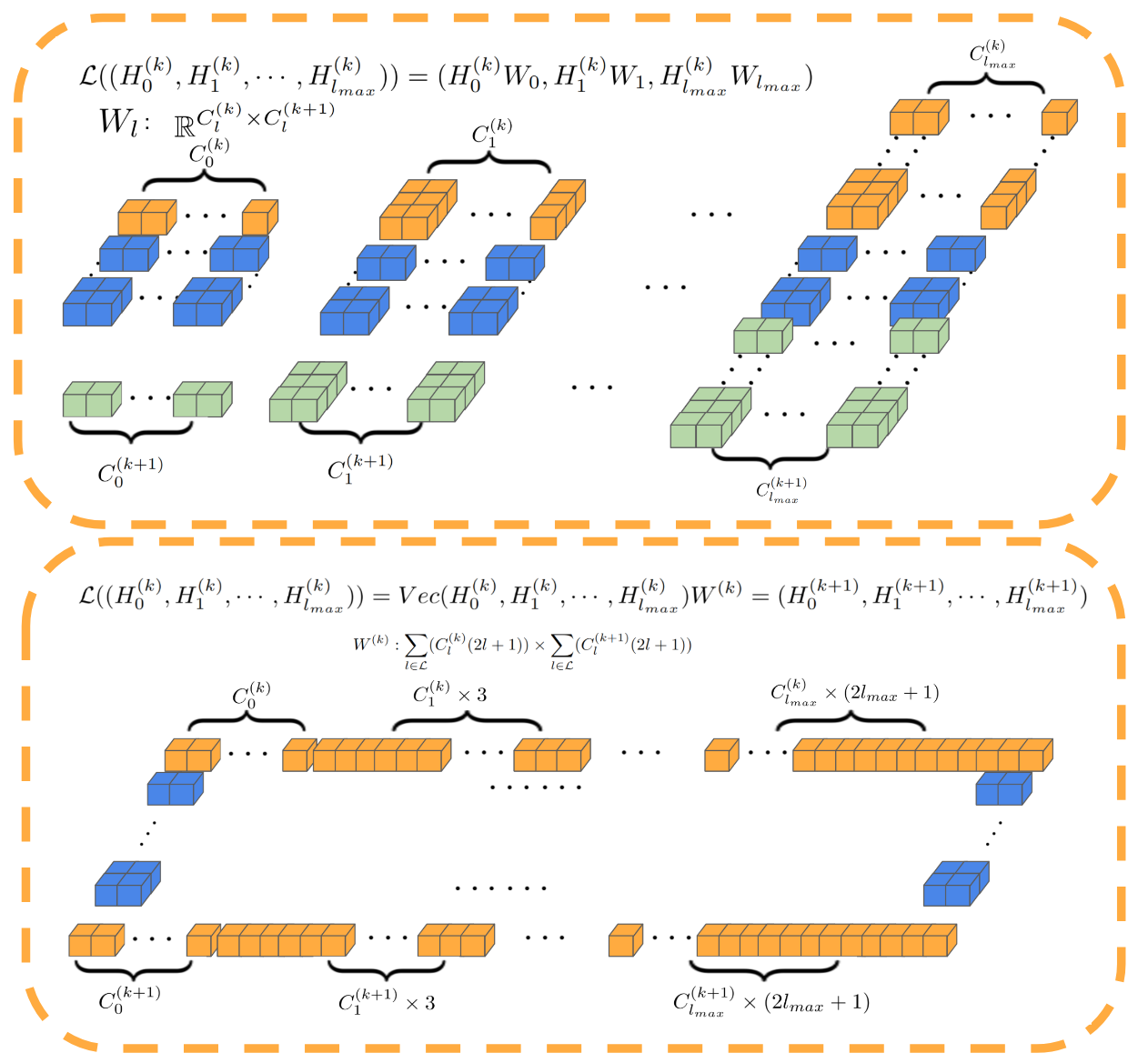}}
\caption{Comparison between Equivariant and Non-Equivariant Linear Layers: The figure above depicts the equivariant linear layer, wherein each type of feature is linearly combined using a specific matrix $W_l$, treating the vector or tensor as a cohesive geometric entity. Conversely, the figure below illustrates the traditional linear approach, where all channels within an intrinsic feature, as well as different types of features, are linearly intermixed, since it vectorizes and concatenates all features and applies a unified weight matrix}
\label{com_linear}
\vskip -0.2in
\end{figure*}

The weights $W_l$ have the format $(C_l^{(k)}, C_l^{(k+1)})$, where $C_l^{(k)}$ is the number of channels in $H^{(k)}_l$, representing the corresponding input channels, and $C_l^{(k+1)}$ is the number of channels in $H^{(k+1)}_l$, representing the corresponding output channels. The difference of equivariant linear layers and conventional linear layers are depicted in Figure \ref{com_linear}.

\vspace{-3mm}

\subsubsection{Equivariant LayerNormalization}

\begin{figure*}
\centering
\vspace{-10mm}
\centerline{\includegraphics[width=6in]{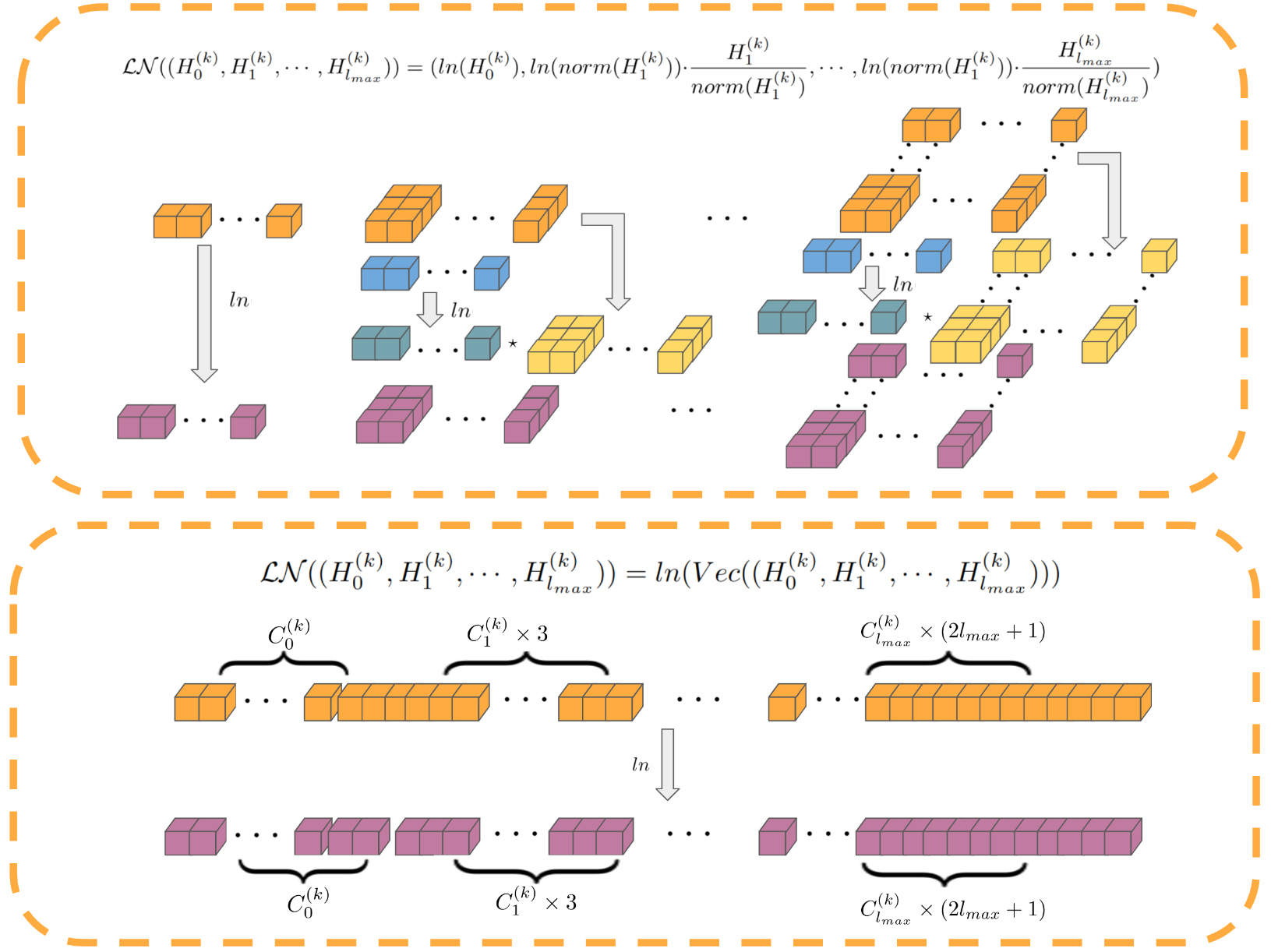}}
\caption{Comparison Between Equivariant and Non-Equivariant Layer Normalization: The figure above illustrates the equivariant nonlinear layer, where layer normalization is applied to the norm of each feature type, followed by multiplication with a unit for each feature. The figure below depicts the traditional nonlinear layer, employing element-wise layer normalization across features, thereby disrupting equivariance by treating each feature type as a concatenation of individual channels rather than a unified geometric entity.}
\label{com_ln}
\end{figure*}

We use the commonly used equivariant layer normalization in equivariant works 
that 
apply the layer normalization to the norm of the features, and then multiply those with unit tensor features. 
The normalization layer $\mathcal{LN}$ is defined as 

\begin{align*}
&\mathcal{LN}((H^{(k)}_0,H^{(k)}_1,\cdots, H^{(k)}_{l_{max}}))\\
&= (ln(H^{(k)}_0), ln(\|H^{(k)}_1\|) \cdot \frac{H^{(k)}_1}{\|H^{(k)}_1\|},\\
&\cdots,ln(\|H^{(k)}_1\|)\cdot \frac{H^{(k)}_{l_{max}}}{\|H^{(k)}_{l_{max}}\|}),
\end{align*}

where $ln$ is the conventional layer normalization. The difference of equivariant layernormalization and conventional layernormalization are depicted in Figure \ref{com_ln}.

\vspace{-3mm}

\subsection{Spherical Harmonics}
\label{spherical_harmonics}
The spherical harmonics constitute a complete set of orthogonal functions, making them an orthonormal basis. Any spherical function $f \in \mathcal{L}^2(\mathbb{S}^2)$ can be expressed as a linear combination of these spherical harmonics. In simpler terms, they serve as the Fourier basis for functions defined on a sphere.

Similar to the varying frequencies of sines and cosines in Fourier series, spherical harmonics  are characterized by different degrees (orders), denoted as $l \in \mathbb{N}$. Each degree of spherical harmonics corresponds to a specific pattern or shape on the surface of a sphere.  Higher orders (degrees) indicate higher frequencies, resulting in more intricate and complex patterns on the sphere's surface. To apply spherical harmonics in three-dimensional space ( $\mathbb{R}^3$), we incorporate a radial component $r^l$ into the original spherical harmonics of the corresponding degree (order)-$l$. This method scales the spherical harmonics for three-dimensional applications, extending their utility beyond the sphere $\mathbb{S}^2$. This adaptation scales the spherical harmonics for applications beyond the two-dimensional sphere surface, effectively extending their utility to three-dimensional analyses and applications. Despite this adaptation, these functions retain their fundamental characteristics and are still referred to as spherical harmonics. In this work, we utilize spherical harmonics, incorporating 3D Cartesian coordinates, as positional embeddings in transformer models. 

The fascinating properties of spherical harmonics make them crucial in $SO(3)$ group representations, allowing us to harness their power to achieve equivariance in transformers. A order-$l$ spherical harmonics, denoted as $Y^l: \mathbb{R}^3 \rightarrow \mathbb{R}^{2l+1}$, is a vector function with $2l+1$ dimension, following the transformation rule:
\begin{align*}
&Y^l(Rr)=D^l(R)Y^l(r),\\
&Y^l(r)=\|r\|^lY^l(\hat{r}),
\end{align*}

where $R$ is an arbitrary rotation, $\hat{r}=\frac{r}{\|r\|}$, and $D^l: SO(3) \rightarrow \mathbb{R}^{(2l+1)\times(2l+1)}$, is called the Wigner-D matrix, serving as the irreducible representation of $SO(3)$ corresponding to the order $l$.  The Wigner-D matrix are orthogonal matrix,  i.e.,  $D^l(R)D^l(R)^T=I$.
%
%
To mitigate the impact of the large scaling factor $\|r\|^l$, we explored alternative approaches: one involved substituting the scaling factor $\|r\|^l$ with Gaussian radial basis functions, expressed as $e^{-(\|r\|-l)^2}$. Another approach entailed employing Fourier sine and cosine series to represent $\|r\|$, creating an invariant embedding. This embedding was then combined with the positional encoding of the unit vector $\hat{r}$ using the spherical harmonics for sphere. However, these alternatives did not demonstrate any significant benefits over the use of spherical harmonics adapted for three-dimensional space ($\mathbb{R}^3$), leading us to continue with our initial methodology.

\subsubsection{The use of Spherical Harmonics in Equivariant Transformer}
We acknowledge that most equivariant transformer works 
\citet{fuchs2020se,liao2022equiformer,liao2023equiformerv2} also uses spherical harmonics in the transformer layers. However, the use of the spherical harmonics is to derive the equivariant kernel basis based on the relitive position for specific geometirc entities. For example, \citet{fuchs2020se} first proposes the $SE(3)$ equivariant transformer for point clouds, where spherical harmonics is served as the equivariant kernel as in steerable 3D convolutions. Equiformer\cite{liao2022equiformer, liao2023equiformerv2} is a graph-based architecture that leverages spherical harmonics for the edges, using a depth-wise tensor product to embed it into the node. For this graph embedding (atom + edge-degree) and graph attention, the use of spherical harmonics could be interpreted more as an equivariant kernel basis. The difference here is that Equiformer uses spherical harmonics to integrate edge information into the equivariant kernel, while ours uses spherical harmonics directly in each token.  

\section{Equivariance of Conventional Positional Encoding}
\label{fourier}
The conventional positional encoding leveraging Fourier sine and cosine functions is translational equivariant. When the input is translated by a translation $t$, the output will be transformed in a specific way, multiplied by the representation of $t$, i.e., $PE(x+t)= e^{i\omega t}e^{i\omega x}$ for specific frequency $\omega$ in complex format or
\begin{align*}
PE(x+t) &= \begin{bmatrix}
cos(\omega t)& -sin(\omega t)\\
sin(\omega t)& cos(\omega t)
\end{bmatrix}  \begin{bmatrix}
cos(\omega x)& -sin(\omega x)\\
sin(\omega x)& cos(\omega x)\\
\end{bmatrix}\\
&=\rho_\omega(t)PE(x)
\end{align*}

in matrix format.

\section{Equivariant Hidden Feature Format}
\label{equi_latent}

\begin{figure}[ht]
\vskip 0.2in
\begin{center}
\centerline{\includegraphics[width=0.7\columnwidth]{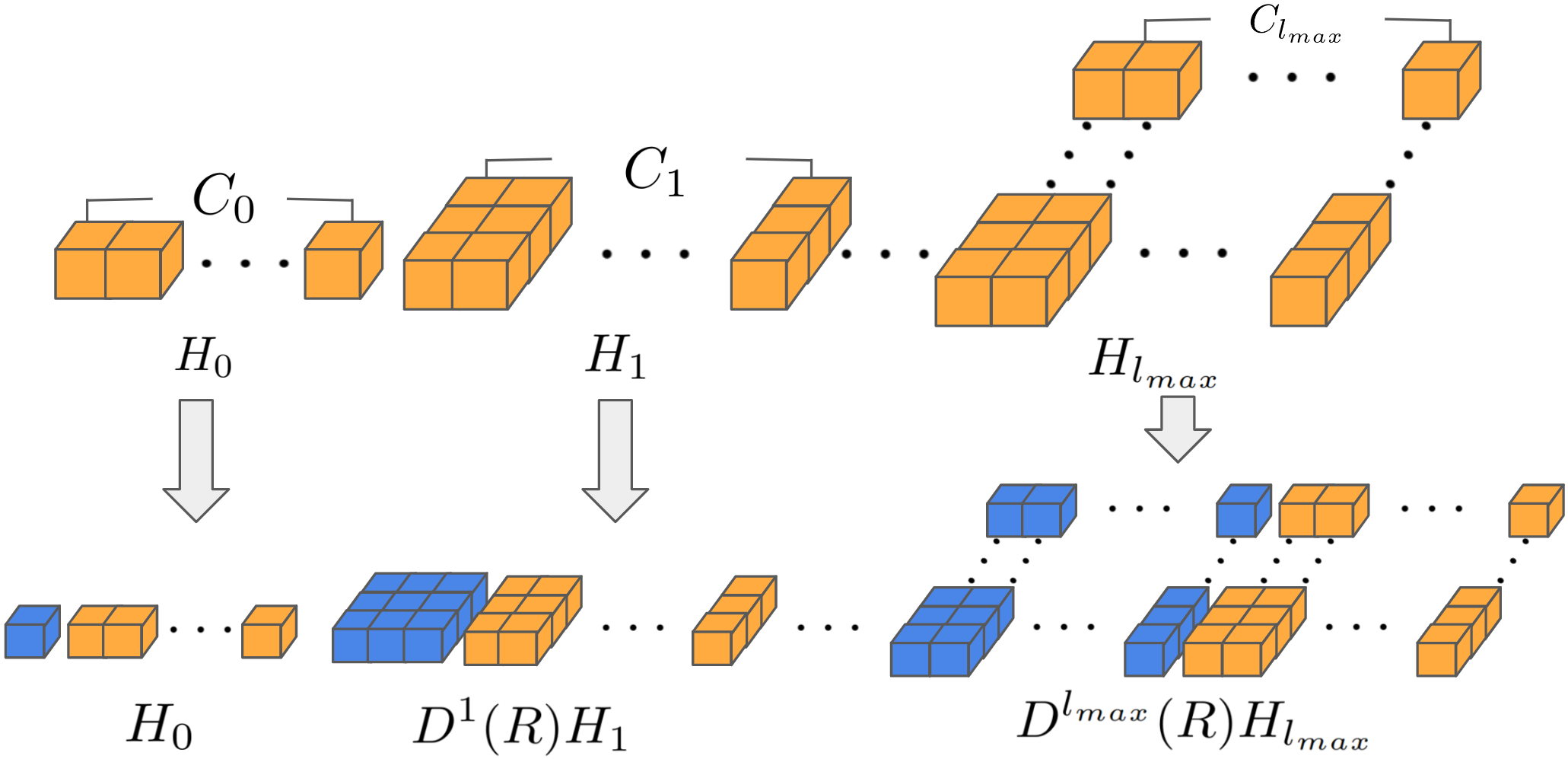}}
\caption{Latent code transformation.}
\label{transformation_of_latent}
\end{center}
\vskip -0.2in
\end{figure}

The equivariant latent code is structured as $H = \bigoplus_{l \in L} H_l = (H_0, H_1, \cdots, H_l)$, with each $H_l$ having dimensions $(2l+1, C_l)$. The action of a rotation $R$ on this latent code is depicted in Figure \ref{transformation_of_latent}.

\section{Visualization of the Equivariant Latent}
\label{visualization_sphere}

With the spherical harmonics, we can introduce the Fourier Transform for the sphere. The Fourier coefficient $\mathcal{F}^l$ of a function on the sphere, $f:\mathbb{S}^2 \rightarrow \mathbb{R}$, corresponding to the order $l$ is obtained by 

\begin{align*}
\mathcal{F}^l = \int_{\mathbb{S}^2}f(x)Y^l(x)dx,
\end{align*}

and the inverse Fourier Transform without normalization follows the equation: 
\begin{align*}
f(x) = \sum_{l} (\mathcal{F}^l)^T Y^l(x)dx,
\end{align*}

When we rotate the function $f$ with any rotation $R \in SO(3)$, i.e.,  we get a new function $f' =f(R^{-1}x)$. The Fourier coefficients corresponding to the order $l$ are: 

\begin{align*}
\mathcal{F}'^l = \int_{\mathbb{S}^2}f(R^{-1}x)Y^l(x)dx = \int_{\mathbb{S}^2}f(y)Y^l(Ry)dy =
\int_{\mathbb{S}^2}f(y)D^l(R)Y^l(y)dy = D^l(R) \mathcal{F}^l,
\end{align*}

This indicates that when a spherical function is rotated by any rotation $R$, its Fourier coefficients will be multiplied by the corresponding Wigner-D matrix. Inversely, we have that when all Fourier coefficients $\mathcal{F}^l$ are multiplied by Wigner-D matrices $D^l(R)$, the obtained spherical function is rotated by $R$. 

With such preliminary, we can treat our latent code $\bigoplus_{l \in L}H_l$ as the Fourier coefficients, where type-$l$ features are the $l$-th order coefficients,  enabling us to derive the spherical function. As a result, when our features are transformed — with each type being multiplied by its respective Wigner-D matrix — the spherical function undergoes a corresponding rotation.

\section{Equivariant Nonlinear Layer}
\label{nonlinearity}

\begin{figure*}
\centering
\vspace{-10mm}
\centerline{\includegraphics[width=6in]{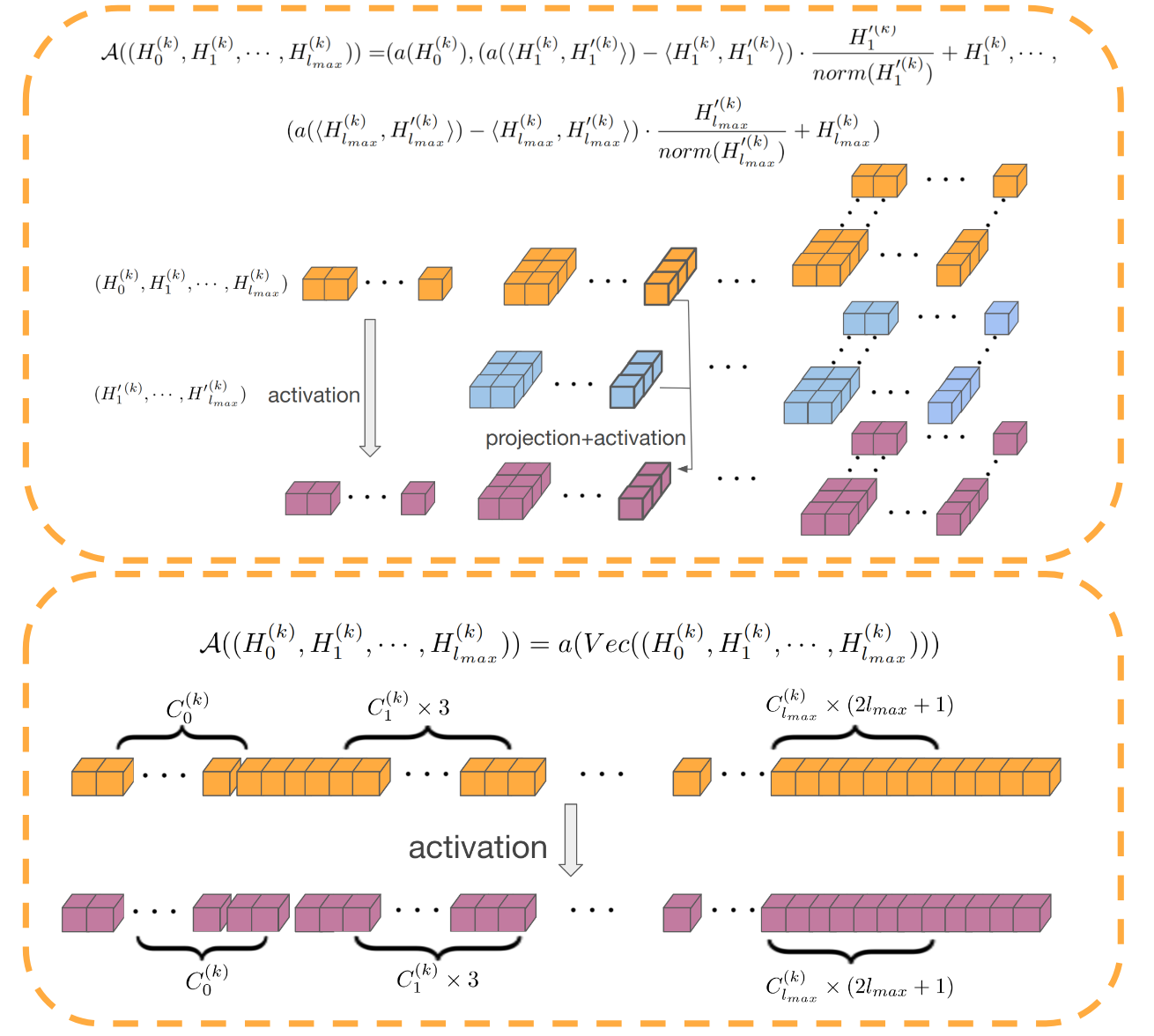}}
\caption{Comparison of Equivariant and Non-Equivariant Nonlinear Layers: The figure above illustrates the equivariant nonlinear layer, similar to those in vector neuron models. This layer establishes equivariant directions using an equivariant linear layer, applies nonlinearity to the projection of the original feature in these directions, and then adds this to the orthogonal component relative to the direction in the input. On the other hand, the figure below shows the conventional nonlinear layer, which applies element-wise nonlinearity to the vectorized feature, thus failing to preserve equivariance. }
\label{com_nl}
\end{figure*}

In our proposed nonlinear layer, we first generate intermediate hidden features $\bigoplus_{l \in L} H'^{k}_l=(H'^{(k)}_1,\cdots, H'^{(k)}_{l_{max}})$ of the same size as the input via an equivariant linear layer. Subsequent to this, we employ the specified nonlinearity:
\begin{align*}
&\mathcal{A}((H^{(k)}_0,H^{(k)}_1,\cdots, H^{(k)}_{l_{max}}))\\
&=(a(H^{(k)}_0), (a( \langle H^{(k)}_1, H'^{(k)}_1\rangle)-\langle H^{(k)}_1, H'^{(k)}_1\rangle) \cdot \frac{H'^{(k)}_1}{norm(H'^{(k)}_1)} + H^{(k)}_1,\\
&\cdots,(a( \langle H^{(k)}_{l_{max}}, H'^{(k)}_{l_{max}}\rangle)-\langle H^{(k)}_{l_{max}}, H'^{(k)}_{l_{max}}\rangle) \cdot \frac{H'^{(k)}_{l_{max}}}{norm(H'^{(k)}_{l_{max}})} + H^{(k)}_{l_{max}})
\end{align*}
Here, $\langle \cdot,\cdot \rangle$ denotes the per-channel inner product, meaning the size of $\langle H^{(k)}_l, H'^{(k)}_l\rangle$ is in the format $(1,C^{(k)}_l)$. The function $a$ represents a conventional activation operation, such as $ReLU$, $Sigmoid$, or $LeakyReLU$. The symbol $\cdot$ indicates broadcast multiplication.
This approach bears resemblance to gated normalization \cite{weiler20183d}. However, in our model, the scalar for the ``gate" is derived from the inner product of two outputs of the equivariant linear layer, rather than being a scalar present in the hidden features themselves. The difference of equivariant nonlinear layers and conventional nonlinear layers are depicted in Figure \ref{com_nl}.

\vspace{-3mm}

\section{Multi-head Attention Inner Product}
\label{App:MH_attention_inner}
Given the input of the attention module formulated as $\bigoplus_{l \in L} H_l$, we generate query $Q$, key $K$, and value $V$ by equivariant linear layers. As $Q$, $K$, and $V$ are equivariant features, they are expressed as $Q_{i}=\bigoplus_{l \in L} (Q_l)_i$, $K_{j}=\bigoplus_{l \in L} (K_l)_j$, and $V_{j}=\bigoplus_{l \in L} (V_l)_j$, where $i$ and $j$ are the indices of the latents. The inner product is calculated between the equivariant key $K$ and the equivariant query $Q$, and here we describe how we use it in multi-head attention. 
With $N_h$ multi-heads, we split the features $K$, $Q$, and $V$ into $N_h$ heads along the channel dimension. Taking $Q$ as an instance for clarity, and denoting $C_l$ as the number of channels for type-$l$ feature $(Q_l)_i$ in $Q_i$, $Q_i$ gets divided into various heads $(Q_i)^h$ with $h$ as the head index. $(Q_i)^h$ maintains the equivariant feature format, represented as $(Q_i)^h = \bigoplus_{l \in L} (Q_l)_i^h$, with the channel count for type-$l$ feature $(Q_l)_i^h$ being $\frac{C_l}{N_h}$. This division also applies to $K$ and $V$. The inner product of $(Q_i)^h$ and $(K_{j})^h$ is defined as
\begin{align}
\langle (Q_i)^h, (K_{j})^h\rangle = \sum_{l \in L} \sum_c^{\frac{C_l}{N_h}}(((Q_l)_i^h)_c)^T ((K_l)_j^h)_c.
\label{inner_product}
\end{align}
With the defined inner product, we have the output:
\begin{align}
(O_i)^h = \sum_j\frac{exp(\langle (Q_i)^h, (K_j)^h\rangle)}{\sum_j exp(\langle (Q_i)^h, (K_j)^h\rangle)}(V_j)^h
\label{multihead_matmul}
\end{align}
The final output of the attention mechanism, composed in the channel dimension, can be denoted as $\bigoplus_{l \in L} O_l$. For proof of equivariance, refer to Sec. \ref{proof_mh}.

\section{Proof of Equivariance for Multi-Head Attention}
\label{proof_mh}
The inner product is formulated as follows:
\begin{align*}
\langle (Q_i)^h, (K_{j})^h\rangle = \sum_{l \in L} \sum_{c=1}^{\frac{C_l}{N_h}}(((Q_l)_i^h)_c)^T ((K_l)_j^h)_c.
\end{align*}
Owing to the equivariance of the Linear Layer, when the input undergoes a rotation $R$, the components $Q, K, V$ are correspondingly transformed, denoted as $R \cdot Q$, $R \cdot K$, and $R \cdot V$. Under these conditions, the corresponding inner product becomes:
\begin{align*}
\langle (R \cdot Q_i)^h, (R \cdot K_{j})^h\rangle &= \sum_{l \in L} \sum_c^{\frac{C_l}{N_h}}((D^l(R)(Q_l)_i^h)_c)^T (D^l(R)(K_l)_j^h)_c\\
&= \sum_{l \in L} \sum_c^{\frac{C_l}{N_h}}(((Q_l)_i^h)_c)^T\bm{D^l(R)^TD^l(R)}((K_l)_j^h)_c\\
&= \sum_{l \in L} \sum_c^{\frac{C_l}{N_h}}(((Q_l)_i^h)_c)^T(K_l)_j^h)_c\\
&= \langle (Q_i)^h, (K_{j})^h\rangle,
\end{align*}
which proves the invariance of the defined inner product. Thereby, the output becomes:

\begin{align*}
&\sum_j\frac{exp(\langle (Q_i)^h, (K_j)^h\rangle)}{\sum_j exp(\langle (Q_i)^h, (K_j)^h\rangle)}(R \cdot V_j^h)\\
&=\sum_j \frac{exp(\langle (Q_i)^h, (K_j)^h\rangle)}{\sum_j exp(\langle (Q_i)^h, (K_j)^h\rangle)} \bigoplus_{l\in L} D^l(R) (V_l)_j^h\\
&=\bigoplus_{l \in L} D^l(R) (\sum_j\frac{exp(\langle (Q_i)^h, (K_j)^h\rangle)}{\sum_j exp(\langle (Q_i)^h, (K_j)^h\rangle)} (V_l)_j^h)\\
&= \bigoplus_{l \in L} D^l(R)(O_l)_j^h\\
& = R \cdot O_j^h,
\end{align*}
which proves that the whole attention mechanism is equivariant.

\section{Alternative Equivariant Attention}
\label{alter_attn}
When it comes to attention mechanisms involving latents with different types of equivariant features, a direct method is to use tensor product to entangle these different types, which is complicated and computationally expensive. However, there is an alternative approach that treats the different types of features as Fourier coefficients to obtain spherical features. By applying the conventional transformer to these spherical features, followed by the Fourier Transform, we can retrieve different types of equivariant features. This method offers a more efficient way to entangle and handle various types of features within the attention mechanism.
For latent features $\bigoplus_{l \in L} H_l$, we apply the Inverse Fourier Transform so that: 
$$S_l(x)= Y^l(x)^T H_l,$$ which implies that $S_l$ has size $(N_R,C_l, N_S)$, where $N_R$ is the number of latents and $N_S$ is the number of samplings for the sphere. From the preliminary in appendix \ref{visualization_sphere}, we know that when the input features are rotated by $R$, the output  becomes $S_l(R^{-1}x)$, which means these spheres are rotated as well. By concatenating the $\left \{ S_l\right\}$ on the channel dimension, we get the features after an inverse Fourier Transform with size $(N_R,\sum_l C_l, N_S)$, i.e., we have $N_R$ spheres with $\sum_l C_l$ channels and $N_S$ number of sampling. We can directly apply the self-attention mechanism to these spheres without breaking equivariance, resulting in spherical features $F$ with dimensions $(N_R, \sum_l C_l, N_S)$ after the self-attention. Finally, we apply the Fourier Transform as follows:
$$H_l = \sum_i S^l(x_i)Y^l(x_i), $$ 
where $S^l(x_i) = F[:,Ind_l,i]$, with $Ind_l$ representing the index of channels for spheres that correspond to type-$l$ features $H_l$ and $i$ denoting the index of the sample on the sphere. 
By composing different types of features, we obtain outputs in the format $\bigoplus_{l \in L} H_l$. It is evident that the composition of inverse Fourier Transform, transformer on the sphere and the Fourier Transform is equivariant, as confirmed by the preliminary properties of the Fourier Transform in appendix \ref{visualization_sphere}.
 In practice, the computational load is increased due to the number of samples and the complexity of spherical convolution. To address this, we can utilize icosahedron sampling and apply equivariant correlation on the icosahedron for the linear layer in attention. Additionally, we can use standard nonlinear layers and typical layer normalization in the self-attention mechanism.

\section{Averaged Global Geometric Embedding}
\label{average_eb}
The formula for the averaged global geometric embedding is as follows:
\begin{align*}
(&\frac{1}{N}\sum_i Y^1(R^1_i)\oplus Y^1(R^2_i)\oplus Y^1(R^3_i), \\
&\frac{1}{N}\sum_i Y^2(R^1_i)\oplus Y^2(R^2_i)\oplus Y^2(R^3_i), \cdots,\\
&\frac{1}{N}\sum_i Y^{l_{max}}(R^1_i)\oplus Y^{l_{max}}(R^2_i)\oplus Y^{l_{max}}(R^3_i)),
\end{align*}
where the superscript denotes the index of the column in the matrix.

\section{Proof of Invariant Latent and Query}
\label{invariant_latent_query}
Given that the predicted frame $R$ is equivariant and the latent code $\bigoplus_{l \in L} (\mathcal{R}_K)_l$, is also equivariant, when the input undergoes a transformation by a rotation $(R_0,t_0) \in SE(3)$, the frame is modified to $R_0R$, and the latent code transforms to $R_0 \cdot \mathcal{R}_K = \bigoplus_{l \in L} D^l(R_0)(\mathcal{R}_K)_l$. Applying the inverse of the equivariant frame to the latent code yields:
\begin{align*}
\bigoplus_{l \in L} D^l(R_0R)^T D^l(R_0)(\mathcal{R}_K)_l &= \bigoplus_{l \in L} D^l(R)^T \bm{D^l(R_0)^TD^l(R_0)}(\mathcal{R}_K)_l = \bigoplus_{l \in L} D^l(R)^T (\mathcal{R}_K)_l,
\end{align*}
demonstrating that the transformed latent code is invariant.
Furthermore, when the input is subjected to a transformation by a rotation $(R_0, t_0) \in SE(3)$, the decoded camera $j$'s pose shifts to $(R_0R_j, R_0t_j+t_0)$. After subtracting the center, the pose becomes $(R_0R_j, R_0(t_j-\bar{t}))$. Applying the inverse of the equivariant frame to the query camera pose, we arrive at:
\begin{align*}
((R_0R)^T(R_0R_j), (R_0R)^T(R_0(t_j-\bar{t}))) &= (R^TR_j, R^T(t_j-\bar{t})),
\end{align*}
which confirms that the transformed query camera pose remains invariant.




\section{Alternative Equivariant Decoder}
\label{sec:alternative}

\begin{figure*}
\centering
\centerline{\includegraphics[width=6in]{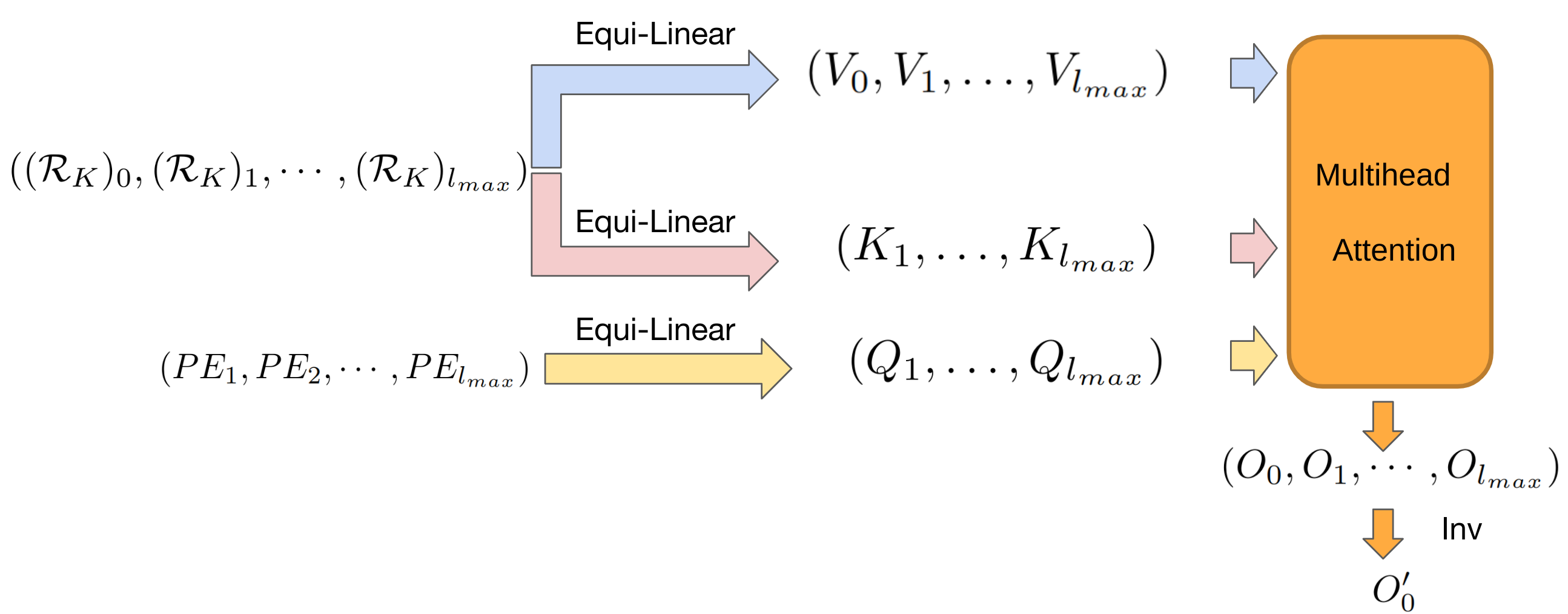}}
\caption{Equivariant Decoder.}
\label{alter_equi_decoder}
\end{figure*}

The pipeline of the equivariant decoder is show in Figure \ref{alter_equi_decoder}. The cross-attention mechanism with equivariance processes two inputs: firstly, the resulting hidden features from self-attention, denoted as $\bigoplus_{l \in L}(\mathcal{R}_K)_l$ with $L=\{0,1,\cdots, l_{max}\}$, and secondly, the positional encoding of query rays and cameras, represented as $(PE(r^j_{uv}, t_j-\bar{t}))$ with spherical harmonics. In this context, $r^j_{uv}$ signifies the $(u,v)$-th ray in the $j$-th query camera, $t_j$ refers to the translation of the $j$-th query camera, and $\bar{t}$ is the pre-calculated center of the encoded cameras. The positional encoding is structured as $\bigoplus_{l \in L}PE_l$ with $L=\{1,\cdots, l_{max}\}$. Note that this encoding does not include the $0$-type (invariant) image features, in contrast to the encoded input.
Hence, when generating the $Q, K, V$ features in the attention module, $Q$ features do not include $0$-th type features. To achieve invariant attention weights, $K$ features should also exclude $0$-th type features, which is accomplished by setting $W_0 = \bm{0}$ in the equivariant linear layer. The multi-head attention mechanism adheres to the equations in Sec. \ref{App:MH_attention_inner}:
\begin{align*}
\langle (Q_i)^h, (K_{j})^h\rangle = \sum_{l \in L} \sum_c^{\frac{C_l}{N_h}}(((Q_l)_i^h)_c)^T ((K_l)_j^h)_c,
\end{align*}
\vspace {-2mm}
\begin{align*}
(O_i)^h = \sum_j\frac{exp(\langle (Q_i)^h, (K_j)^h\rangle)}{\sum_j exp(\langle (Q_i)^h, (K_j)^h\rangle)}(V_j)^h,
\end{align*}
where $L =\{ 1,2,\cdots, l_{max}\}$, and $(V_j)^h$ conforms to the $\bigoplus_{l \in \{0,1, \cdots, l_{max}\}}(V_l)_j^h$ format. The output $O$ is derived as $\bigoplus_{l \in \{0,1, \cdots, l_{max}\}} O_l$.

\begin{figure*}
\centering
\centerline{\includegraphics[width=6in]{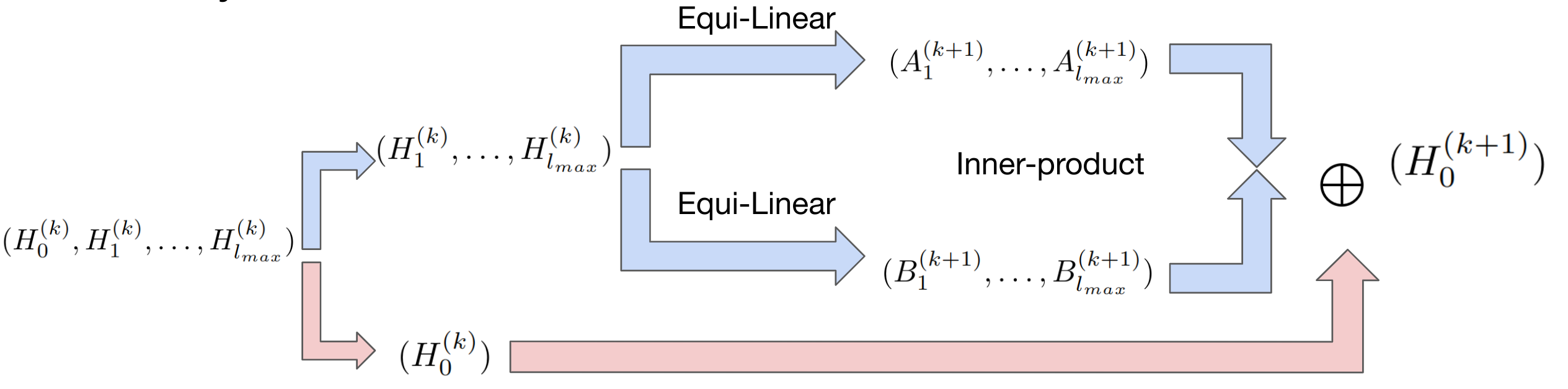}}
\caption{Invariant Layer}
\label{inv_layer}
\end{figure*}

Since the final prediction value is unchanged when the reference frame is transformed, it is characterized as an invariant type-$0$ (scalar) feature. To extract the invariant features for prediction, we utilize an ``invariant layer", as depicted in Figure \ref{inv_layer}. In this process, we initially generate two intermediate features, $H'$ and $H''$, of the same size. Subsequently, for each type-$l$, we perform an inner product operation between $H'_l$ and $H''_l$, which results in invariant features $I_l$, each with a channel count of $C_l$. By concatenating these invariant features $\{I_l\}$, we formulate final invariant features $ O'_0 = I $  with a combined channel count of $\sum_l C_l$, which is then utilized for the final prediction.

\section{Network Architecture and Implementation Details}
\label{App:Network_Arch}
Regarding architecture, we use a ResNet18 as the visual backbone, resulting in 960-dimensional features. The order of spherical harmonics is [1,2,4,8], resulting in (3+5+9+17)*2 = 68-dimensional features. For encoding, visual and geometric features are concatenated to produce 960 + 68 = 1028-dimensional embeddings. For decoding, we use the same Fourier encoding as the standard Perceiver IO, resulting in 186-dimensional embeddings. Our original latent representation R is of dimensionality 1024 x 512. We set the number of channels for each type of the equivariant hidden feature as [512, 64,32, 8]. In the Perceiver IO implementation, we have 1 block of cross-attention with 1 head, 8 self-attention layers with 8 heads, and 1 cross-attention with 1 head for a decoder.

Our DeFiNe baseline has ~73M parameters, and our EPIO implementation has ~147M parameters. This increase is due to: additional parameters for the global geometric latent code as shown in Figure 5; inference for frame prediction as shown in Figure 2; and additional parameters for type-2, type-3, and type-4 features, where we set the channel numbers as 64, 32, and 8 respectively. Regarding runtime, we observed an increase of roughly 2x in training iteration times and 1.5x in per-pixel queries during inference. However, we would like to note that our approach converges in roughly 15\%. Training and evaluation was conducted using distributed training (DDP) on 8 A100 GPUs, with 80 GB each. 

Regarding experiments, we used Pytorch to implement our Equivariant Perceiver IO and will open-source our code and pre-trained weights upon acceptance. We used a batch size of 192, the AdamW optimizer with $\beta = 0.9$, and $\beta_2 =0.999$, weight decay of $10^{-4}$, and an initial learning rate lr at $2 \times 10^{-4}$.
. For ScanNet, the training duration was 200 epochs, with the learning rate being reduced by half every 80 epochs; For DeMon datasets, the training duration was 200 epochs, with the learning rate being reduced by half every 80 epochs. We used the same losses as DeFiNe, i.e., the L1-log loss, with a weight of 1.0 for real views and 0.2 for virtual views. Following standard practice, we used images of size 128x192 for ScanNet, and images of size 240x320 for DeMoN, using two images as input, with corresponding intrinsics and extrinsics, and ground-truth depth maps as supervision (see Section 3.1). This is the standard training and evaluation protocol and was used by our baselines as well, ensuring a fair comparison.

\section{Extended Discussion for General Tasks of Equivariant Periceiver IO}
\label{general_task}

\input{task_table}

Our model, designed as a general architecture, is adaptable to various tasks. This paper focuses on demonstrating the advantages of integrating equivariance into Perceiver IO for scene representation, primarily evaluated through depth estimation, a core problem in vision. While implementing our model for other tasks is beyond this paper's scope, we provide a brief overview of its potential extensions to different applications, as shown in table \ref{task_table_ex}. 

\begin{figure*}
\centering
\vspace{-10mm}
\centerline{\includegraphics[width=6in]{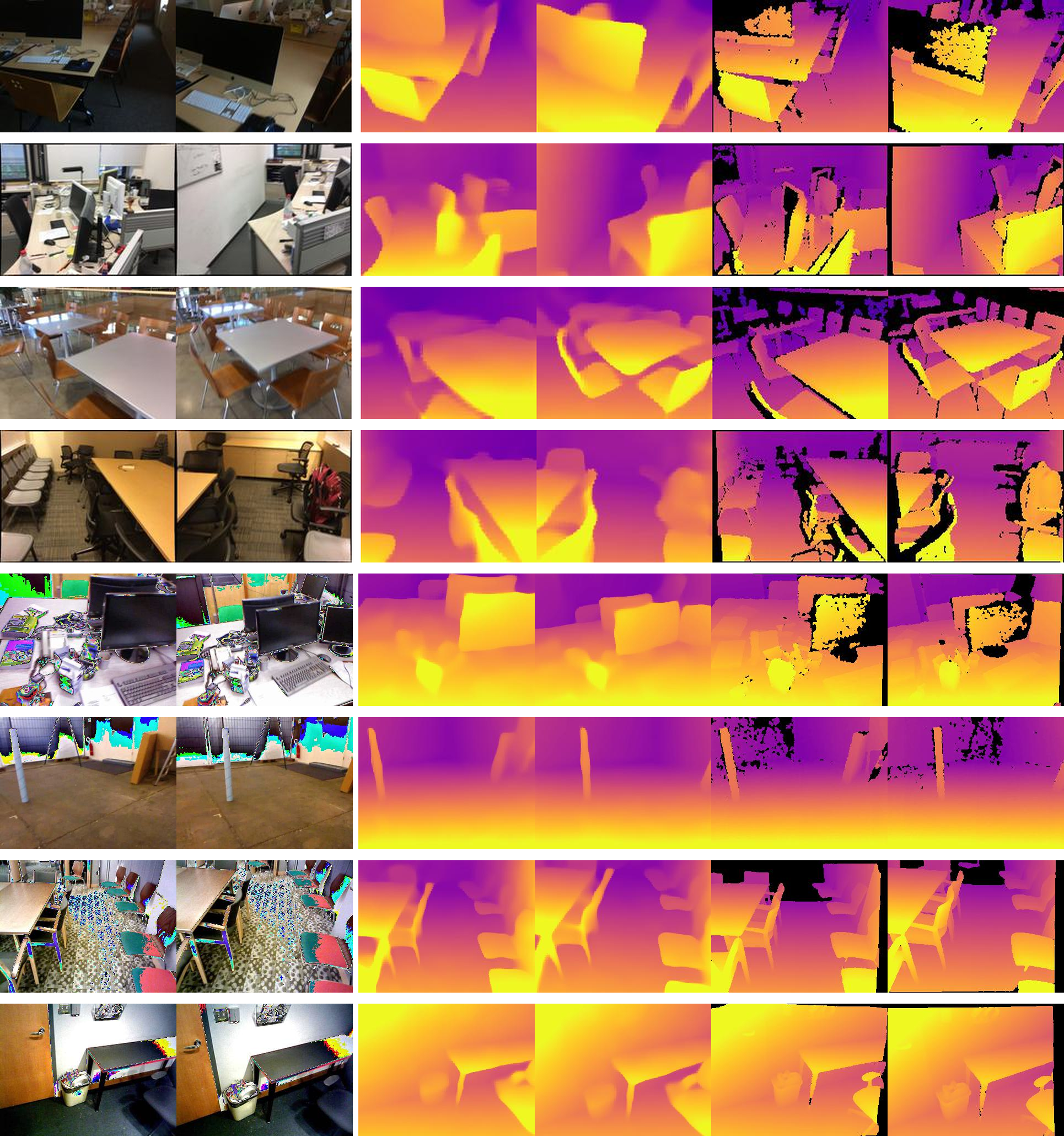}}
\caption{Qualitative Results}
\label{qualitative_results_temp}
\end{figure*}

\section{More Qualitative Results for Depth Estimation}
Please see Figure \ref{qualitative_results_temp} for more qualitative results.

\section{More Experiments}

\subsection{Comparison with Current Prevalent Depth Estimation Model}
\label{depth_anything_app}

We have evaluated DepthAnything on the same ScanNet stereo benchmark we report results as shown in Table\ref{tab:depthanything}. we can confidently say that our method outperforms DepthAnything on this benchmark. However, we would like to emphasize that these are not meaningful comparisons. DepthAnything is a monocular depth estimation network that outputs affine-invariant predictions, while ours is a multi-view depth estimation network that outputs metric predictions. Hence, to achieve the reported DepthAnything numbers, we had to artificially shift and scale predictions using ground-truth depth maps (the same thing is done in their paper). We also could not use the second image as input to DepthAnything, since it is a monocular network, while our method can leverage multiple images as input by design (and even benefits from that during training via the virtual camera augmentation procedure).

\begin{table}[t]
\renewcommand{\arraystretch}{0.95}
\vskip -0.1in
\begin{center}
\begin{small}
\begin{tabular}{lcccr}
\toprule
Models & Abs.Rel.$\downarrow$ & RMSE$\downarrow$ & $\delta < 1.25\uparrow$ \\
\midrule
Depth anything& 0.099&0.226& 0.903\\
\midrule
Ours& \textbf{0.076}&\textbf{0.217}& \textbf{0.934}\\
\bottomrule
\end{tabular}
\caption{Comparison of our model and Depth Anything}
\label{tab:depthanything}
\end{small}
\end{center}
\end{table}

\subsection{Varying views}
\label{varying_view}

We performed additional small-scale experiments to study the impact of the number of available views. In this setting, we have 500 views of one scene, and we randomly choose N encoding views and 1 different decoding camera viewpoint for novel depth estimation.

For DeFiNe, we train with jittering augmentation on the reference frames and test with augmentation as well. For our model, we train without jittering augmentation and also test with augmentation. We explored 2, 3, and 4 views, and the Abs. Rel. depth estimation results are reported in Tab. \ref{tab:ablation_2}

\begin{table}[t]
\renewcommand{\arraystretch}{0.95}
\begin{center}
\begin{small}
\begin{tabular}{lcccr}
\toprule
 & 2 views & 3 views& 4 views\\
\midrule
DeFiNe  & 0.324&0.315& 0.307\\
Ours & 0.215&0.209& 0.198\\
\bottomrule
\end{tabular}
\caption{Novel View Depth Estimation across a varying number of views.}
\label{tab:ablation_2}
\end{small}
\end{center}
\end{table}

As we can see, our method consistently surpasses DeFiNe across a varying number of views, even without employing augmentation during training, with the performance gap remaining similar across different view counts.

\section{Impact Statements}
\label{App:social_impacts}
This work aims to advance $3D$ effective learning, with an application in $3D$ reconstruction. While the direct outcome of our research may not have immediate societal implications, the broader application of 3D reconstruction technologies could have some impact. A primary concern is the potential for privacy violations, particularly in scenarios where 3D reconstruction is used to create detailed representations of real-world environments or individuals without their consent. Such applications could lead to unauthorized surveillance or data collection, posing ethical and privacy challenges that need to be addressed as this technology advances and becomes more accessible.

%% file: task_table.tex
\begin{table}[t]
\setlength{\tabcolsep}{0.2em}
\begin{center}
\begin{small}
\begin{tabular}{lccccccr}
\toprule
Task & Input & Transformation &Pos. Encoding & Feature Embedding &Query&Prediction \\
\midrule
Novel View Synthesis
& Image + Camera &$SE(3)$ & SPH & Image embedding& camera pose & RGB\\
Neural Volume Rendering  
&Image + Camera & $SE(3)$ & SPH & Image embedding & point+ray dir& $\sigma$, RGB \\
Pose Estimation
&Image + Camera & $SE(3)$ & SPH & Image embedding& image (Inv)& $R,t$\\
Implicit Field for PC
& Point &$SE(3)$ &SPH & Point embedding/-& point &Field value\\
$2D$ Dense Prediction
& Image + Pixel & SE(2) & Trig & $SO(2)$-equi feature& pixel & Field value\\
\bottomrule
\end{tabular}
\caption{Different tasks and their corresponding geometric information in Equivariant Perceiver IO}
\label{task_table_ex}
\end{small}
\end{center}
\end{table}